\documentclass[sigconf]{acmart}
\AtBeginDocument{%
  }

\setcopyright{acmlicensed}
\copyrightyear{2025}
\acmYear{2025}
\acmDOI{XXXXXXX.XXXXXXX}
\acmConference[MM '25]{the 33rd ACM
International Conference on Multimedia}{October 27--31,
  2025}{Dublin, IreLand}
\acmISBN{978-1-4503-XXXX-X/2025/10}

\newcommand*\rotvtwo{\rotatebox{90}}
\usepackage{colortbl}
\usepackage{amsmath}
\usepackage{amsfonts}
\usepackage{booktabs} 

\usepackage{multirow}
\usepackage{fancyvrb}
\usepackage{arydshln}

\usepackage{graphicx}
\usepackage{subcaption}
\usepackage{wrapfig,lipsum,booktabs}
\usepackage{float}

\usepackage{spverbatim}

\usepackage{tcolorbox}

\usepackage{amsthm}
\theoremstyle{plain}

\newtheorem*{thm*}{Theorem}
\theoremstyle{definition}

\newcommand{\rowcolorbase}{\rowcolor{gray!20}}

\usepackage{pgfmath}

\usepackage{xcolor}
\newlength\MAX 
\newlength{\tempLength}
\def\plusbarC#1#2{%
  \pgfmathsetlength{\tempLength}{#2/#1 * \MAX}
  {\color{black!60}\rule{0.001mm}{2ex}}{\color[rgb]{0,0.65,0.35}\rule{\tempLength}{2ex}}{\color{black!20}\rule{\dimexpr\MAX-\tempLength}{2ex}}
}

\def\plusbarB#1#2{%
  \pgfmathsetlength{\tempLength}{#2/#1 * \MAX}
  {\color{black!60}\rule{0.001mm}{2ex}}{\color[rgb]{1,0.549,0}\rule{\tempLength}{2ex}}{\color{black!20}\rule{\dimexpr\MAX-\tempLength}{2ex}}
}

\def\plusbar#1#2{%
  \pgfmathsetlength{\tempLength}{#2/#1 * \MAX}
  {\color{black!60}\rule{0.001mm}{2ex}}{\color[rgb]{0,0.447,0.741}\rule{\tempLength}{2ex}}{\color{black!20}\rule{\dimexpr\MAX-\tempLength}{2ex}}
}

\newcommand{\argmax}{\mathop{\rm arg~max}\limits}
\newcommand{\argmin}{\mathop{\rm arg~min}\limits}
\begin{document}

\title{Quality Text, Robust Vision: The Role of Language in Enhancing Visual Robustness of Vision-Language Models}

\author{Futa Waseda}
\affiliation{%
  \institution{The University of Tokyo}
  \state{Tokyo}
  \country{Japan}
}
\email{futa-waseda@g.ecc.u-tokyo.ac.jp}

\author{Saku Sugawara}
\affiliation{%
  \institution{National Institute of Informatics}
  \state{Tokyo}
  \country{Japan}
}
\email{saku@nii.ac.jp}

\author{Isao Echizen}
\affiliation{%
  \institution{The University of Tokyo, National Institute of Informatics}
  \city{Tokyo}
  \country{Japan}
}
\email{iechizen@nii.ac.jp}







\renewcommand{\shortauthors}{Waseda et al.}

\begin{abstract}
Defending pre-trained vision-language models (VLMs), such as CLIP, against adversarial attacks is crucial, as these models are widely used in diverse zero-shot tasks, including image classification.
However, existing adversarial training (AT) methods for robust fine-tuning largely overlook the role of language in enhancing visual robustness. Specifically, (1) supervised AT methods rely on short texts (e.g., class labels) to generate adversarial perturbations, leading to overfitting to object classes in the training data, and (2) unsupervised AT avoids this overfitting but remains suboptimal against practical text-guided adversarial attacks due to its lack of semantic guidance.
To address these limitations, we propose \textbf{Quality Text-guided Adversarial Fine-Tuning (QT-AFT)}, which leverages high-quality captions during training to guide adversarial examples away from diverse semantics present in images.
This enables the visual encoder to robustly recognize a broader range of image features even under adversarial noise, thereby enhancing robustness across diverse downstream tasks.
QT-AFT overcomes the key weaknesses of prior methods---overfitting in supervised AT and lack of semantic awareness in unsupervised AT---achieving state-of-the-art zero-shot adversarial robustness and clean accuracy, evaluated across 16 zero-shot datasets.
Furthermore, our comprehensive study uncovers several key insights into the role of language in enhancing vision robustness; for example, describing object properties in addition to object names further enhances zero-shot robustness.
Our findings point to an urgent direction for future work---centering high-quality linguistic supervision in robust visual representation learning.

\end{abstract}




\begin{teaserfigure}
  \includegraphics[width=\textwidth]{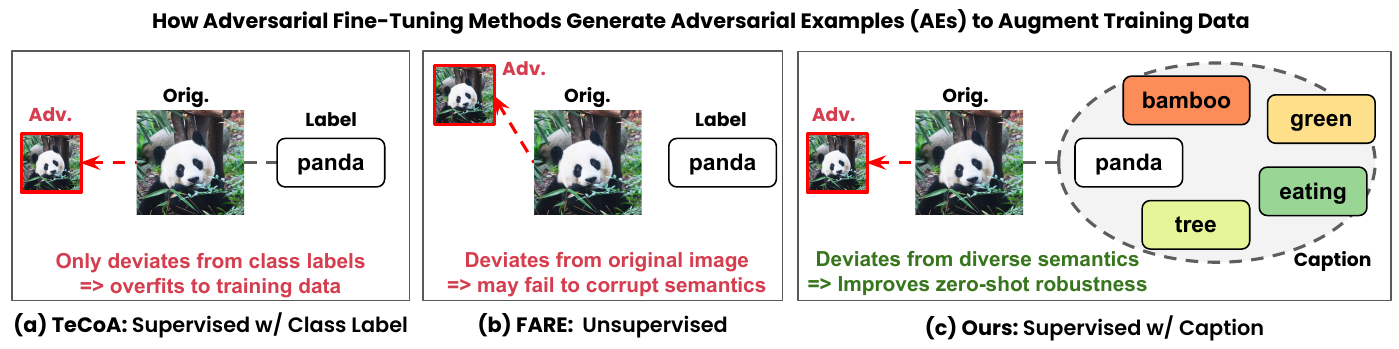}
  \caption{Illustration of how adversarial fine-tuning methods generate adversarial examples (AEs) to augment training data for enhancing zero-shot robustness.
  TeCoA~\cite{mao2022understanding} targets deviation from class labels, while FARE~\cite{schlarmann2024robust} focuses on deviation from the original image. 
  In contrast, our method maximizes deviation from image captions, encouraging divergence from the diverse semantics present in images. This encourages robustness across diverse downstream zero-shot tasks.}
  \label{fig:teaser}
\vspace{0.5cm}
\end{teaserfigure}


\maketitle



\section{Introduction}
Pre-trained vision-language (VL) models, such as CLIP~\cite{radford2021learning}, are trained on large-scale image-text pairs via contrastive learning, enabling the models to obtain joint image-text representations.
This approach allows them to perform a variety of zero-shot tasks, such as zero-shot image classification, where images are matched with arbitrary class labels by comparing image embeddings with the text embeddings of those labels (e.g., ``a photo of \{\textit{class}\}'').
However, recent studies reveal that CLIP is vulnerable to adversarial examples (AEs)~\citep{mao2022understanding, schlarmann2024robust}, which introduce imperceptible perturbations on input images, leading to incorrect model predictions.
This vulnerability poses significant risks in real-world applications. 
Given the widespread adoption of VL models like CLIP, ensuring zero-shot robustness is a critical challenge in building reliable AI systems.

To address adversarial vulnerability, recent studies~\cite{mao2022understanding, wang2024pre, schlarmann2024robust} have proposed robust fine-tuning methods for CLIP's vision encoder based on adversarial training (AT)~\cite{madry2017towards}.
These approaches achieve robustness by fine-tuning for only a few epochs rather than performing AT from scratch, making them more practical.
Additionally, they focus on enhancing zero-shot robustness by assuming that downstream tasks are unknown during fine-tuning and aiming to generalize robustness across diverse zero-shot datasets.

However, we point out that existing defense methods largely overlook the role of language in enhancing vision robustness, making them suboptimal for achieving zero-shot robustness (Fig.~\ref{fig:teaser}). 
For example, supervised (text-guided) AT methods, such as TeCoA~\cite{mao2022understanding}, PMG-AFT~\cite{wang2024pre}, and TGA-ZSR~\cite{yu2025text} rely solely on class labels to guide adversarial perturbations during training (Fig.~\ref{fig:teaser}a). 
By depending on class labels, these methods are highly prone to overfitting on the trained dataset, limiting generalization to unseen downstream tasks.
In contrast, FARE~\cite{schlarmann2024robust} employs an unsupervised AT approach that avoids text guidance, mitigating overfitting (Fig.~\ref{fig:teaser}b). However, due to the absence of semantic guidance from texts, it may fail to capture the diverse semantics present in images during training, limiting its robustness in a wide range of downstream tasks that involve diverse objects or image properties.



To address these challenges, this work introduces a novel perspective on the importance of leveraging language for robust vision in VL models.
Specifically, we propose a simple yet highly effective approach---\textbf{Quality Text-guided Adversarial Fine-Tuning (QT-AFT)}---which leverages detailed image captions instead of simple class labels to enhance the zero-shot robustness of CLIP (Fig.~\ref{fig:teaser}c). 
By incorporating detailed descriptions, the visual encoder learns to robustly recognize a broader range of image features even under adversarial noise, thereby improving performance on diverse downstream tasks.
This approach contrasts with existing text-guided AT methods, which use simple text embeddings of ``a photo of \{\textit{class}\}'' for image classification. 

We conduct extensive experiments by training CLIP on ImageNet and evaluating it across 16 zero-shot datasets. The results show that our method significantly enhances robustness, achieving state-of-the-art zero-shot robustness on 12 out of the 16 datasets and the best average performance. 
Moreover, unlike existing supervised AT methods, our approach does not sacrifice accuracy on clean images; instead, it maintains state-of-the-art accuracy.
These findings highlight that our approach effectively addresses the overfitting issues in supervised AT and the lack of semantic awareness in unsupervised AT.

Furthermore, our comprehensive study uncovers several key insights into the role of language in enhancing vision robustness.
For example, we demonstrate that describing object properties using adjectives and adverbs---not just mentioning objects---further enhances zero-shot robustness.
Additionally, for texture classification tasks where class labels describe textures using adjectives, removing nouns from captions can further improve robustness, showing that the effectiveness of language guidance is task-specific.

By highlighting the critical role of language in enhancing visual robustness, our work points to an urgent direction for future work---centering high quality linguistic supervision in robust visual representation learning.
This direction is unique to multimodal models and distinguishes itself from a wide range of studies focused on unimodal AT methods for traditional image classification tasks.

Our contributions are summarized as follows:
\begin{itemize}
    \item We highlight that existing adversarial fine-tuning methods for CLIP overlook the critical role of language in enhancing the visual robustness of VL models.
    \item We propose \textbf{Quality Text-guided Adversarial Fine-Tuning (QT-AFT)}, which leverages detailed image captions to guide adversarial training. QT-AFT enables the visual encoder to recognize diverse features under adversarial noise, achieving state-of-the-art robustness while maintaining high clean accuracy across downstream tasks.
    \item Our analysis provides key insights into the role of language in enhancing vision robustness, showing that linguistic cues---such as describing object properties in addition to object names---further enhances zero-shot robustness.

\end{itemize}



\section{Related Work}

\paragraph{\textbf{Adversarial Robustness.}}
Adversarial attacks and defenses has been studied extensively in the context of image classification~\citep{szegedy2013intriguing, goodfellow2014explaining_harnessing_ae}.
Adversarial attacks introduce slight perturbations to the inputs to mislead the models' predictions, while maintaining imperceptibility to humans. This poses significant risks of causing unintended consequences in real-world applications of computer vision models.
To mitigate this issue, the defacto standard defense strategy against adversarial attacks is adversarial training (AT)~\citep{madry2017towards}, which augments the training data with AEs to improve model robustness.

\paragraph{\textbf{Adversarial Defense for Vision-Language Models.}}
Many recent vision-language (VL) models~\cite{li2021align, yang2022vision, alayrac2022flamingo, li2023blip} are fundamentally based on CLIP, which learns joint image-text representations by training on a large scale image-text pairs using multimodal contrastive learning.
As a result, existing defense strategies~\cite{mao2022understanding, wang2024pre, yu2025text, schlarmann2024robust} for VL models focus on the CLIP model and perform adversarial fine-tuning on the pre-trained CLIP. 
\citet{mao2022understanding} first proposed novel problem settings of zero-shot robustness in image classification tasks, where the CLIP model must robustly recognize images under adversarial perturbations on unseen downstream datasets. They introduced the first adversarial fine-tuning method for CLIP, called TeCoA, which conducts text-guided contrastive AT by leveraging text embeddings of class labels to obtain robust vision encoder. 
Subsequently, PMG-AFT~\cite{wang2024pre} improved TeCoA by incorporating guidance from a pre-trained model, and TGA-ZSR~\cite{yu2025text} further enhanced robustness by introducing an attention-guided mechanism.
However, these supervised methods based on the class labels tend to overfit to the training dataset, making them suboptimal for achieving zero-shot robustness.
In contrast, FARE~\cite{schlarmann2024robust} proposes an unsupervised AT mechanism that does not rely on text embeddings, thus avoiding overfitting. However, unsupervised AT is also suboptimal due to the text-guided nature of adversarial attacks in practical attack scenarios.

Our method distinguishes itself from both supervised AT methods using class labels and unsupervised AT methods. We introduce a novel supervised AT approach that leverages high-quality captions to guide adversarial perturbations during training.

\section{Methodology}

In this section, we first introduce our problem setup and provide necessary background in Sec.~\ref{subsec:preliminaries}. Next, we analyze the adversarial attack strategies employed in existing adversarial fine-tuning methods, and highlight their limitations in  Sec.~\ref{subsec:analysis}. Finally, in Sec.~\ref{subsec:qt-aft}, we present our proposed method---\textbf{Quality Text-guided Adversarial Fine-Tuning (QT-AFT)}---which addresses these limitations and improves zero-shot robustness.

\subsection{Preliminaries}
\label{subsec:preliminaries}

Following recent efforts to enhance adversarial robustness of VL models~\cite{mao2022understanding, wang2024pre, yu2025text, schlarmann2024robust}, this work focuses on robustly fine-tuning CLIP, the most fundamental and widely used VL model.

\paragraph{\textbf{Vision-Language Contrastive Learning.}}
CLIP consists of an image encoder $f_{\theta}: \mathbb{R}^{d_{\text{I}}} \to \mathbb{R}^{d_\text{E}}$ and a text encoder $f_{\phi}: \mathbb{R}^{d_{\text{T}}} \to \mathbb{R}^{d_{E}}$, where $\theta$ and $\phi$ are their respective parameters, $d_{\text{I}}$ and $d_{\text{T}}$ are the input dimensions of image and text, and $d_\text{E}$ is the joint embedding dimension.
Given an image $x \in \mathbb{R}^{d_{\text{I}}}$ and a text $t \in \mathbb{R}^{d_{\text{T}}}$, CLIP is trained to project them into a shared embedding space, maximizing the cosine similarity of image-text embeddings $\cos(f_{\theta}(x), f_{\phi}(t))$ for correct image-text pairs while minimizing it for incorrect pairs.
CLIP is trained using the InfoNCE loss on a batch of $N$ image-text pairs $\{(x_i, t_i)\}_{i=1}^N$. The InfoNCE loss over images is formalized as:
\begin{equation}
    \label{eq:clip_loss}
    \mathcal{L}_{\text{CLIP-I}}(x, t) = - \sum_{i=1}^N \log \frac{\exp(\cos(f_{\theta}(x_i), f_{\phi}(t_i))/\tau)}{\Sigma_{j=1}^N \exp(\cos(f_{\theta}(x_i), f_{\phi}(t_j))/\tau)},
\end{equation}
where $\tau$ is the learnable temperature parameter. 
The overall loss is the average of the image-to-text and text-to-image losses, given by $\mathcal{L}_{\text{CLIP}} = (\mathcal{L}_{\text{CLIP-I}} + \mathcal{L}_{\text{CLIP-T}}) / 2$, where $\mathcal{L}_{\text{CLIP-T}}$ is the InfoNCE loss over texts.

\paragraph{\textbf{Zero-shot Robustness in Image Classification}}
Using the joint embedding space of the image and text, CLIP is capable of zero-shot image classification.
Given a set of $K$ class templates $c_k$ (e.g., ``a photo of \{\textit{class}\}''), CLIP compares the image embedding with text embeddings and selects the class with the highest similarity:
\begin{align}
    \argmax_{k=1,...,K} \cos(f_{\theta}(x), f_{\phi}(c_k)).
\end{align}
Text embeddings can be created for arbitrary class names, allowing CLIP to perform classification on diverse datasets without additional training—that is, in a zero-shot manner.

However, CLIP's zero-shot classification is vulnerable to adversarial attacks~\cite{mao2022understanding}, where small, imperceptible perturbations to the input image can significantly alter the model's prediction.
Given an image $x$ with true label $y \in {1, ...,K}$, an AE $x'$ is crafted to satisfy:
\begin{align}
    \argmax_{k=1,...,K} \cos(f_{\theta}(x'), f_{\phi}(c_k)) \neq y.
\end{align}
Such AEs can be generated using methods like Projected Gradient Descent (PGD)~\cite{mao2022understanding}, which iteratively perturbs the image to maximize classification loss while constraining the perturbation within an $\ell_p$-norm ball, i.e., $|x' - x|_p < \epsilon$, where $\epsilon$ controls the maximum perturbation size.

To address this vulnerability, we aim to enhance the zero-shot robustness of CLIP through adversarial fine-tuning. Specifically, we adversarially fine-tune the pre-trained CLIP model on a target dataset, such as ImageNet~\cite{deng2009imagenet}, and subsequently evaluate its zero-shot robustness against AEs across diverse unseen datasets.

\paragraph{\textbf{Supervised Adversarial Fine-Tuning for CLIP}}
Supervised adversarial fine-tuning methods for CLIP leverage text embeddings—specifically, class templates $c_k$—as guidance during training. This line of work was initiated by TeCoA, which fine-tunes the vision encoder $\theta$ by minimizing the classification loss on AEs. The objective is formulated as:
\begin{align}
    \mathcal{L}_{\text{TeCoA}}(x,y) = - \log \left( \frac{\exp(\cos(f_{\theta}(x), f_{\phi}(c_y)))}{\sum^K_{k=1} \exp(\cos(f_{\theta}(x), f_{\phi}(c_k)))} \right),  \\
     \theta = \argmin_{\theta} \mathbb{E}_{(x,y) \sim \mathcal{D}} \left[ \max_{x' \in B(x,\epsilon)}  \mathcal{L}_{\text{TeCoA}}(x',y) \right] ,
 \label{eq:tecoa}
\end{align}
where $(x, y)$ is sampled from the data distribution $\mathcal{D}$, $x'$ is the AE generated from $x$, and $B(x, \epsilon)$ denotes the allowed adversarial region (e.g., an $\ell_p$-norm ball).
Here, an AE $x'$ is generated to maximize the cross-entropy loss, while the model parameters are optimized to minimize it.
PMG-AFT~\cite{wang2024pre} and TGA-ZSR~\cite{yu2025text} build upon TeCoA by introducing additional loss functions to further enhance its adversarial robustness.

\paragraph{\textbf{Unsupervised Adversarial Fine-Tuning for CLIP}}
FARE~\cite{schlarmann2024robust} employs unsupervised adversarial fine-tuning to mitigate the overfitting issue observed in TeCoA, avoiding reliance on the text encoder. Specifically, FARE optimizes the following objective:
\begin{align}
     \theta = \argmin_{\theta} \mathbb{E}_{(x,y) \sim \mathcal{D}} \left[ \max_{x' \in B(x,\epsilon)}  ||f_\theta(x') - f_{{\theta}_{\text{orig}}}(x) ||^2_2\right] ,
 \label{eq:fare}
\end{align}
where ${\theta}_{\text{orig}}$ denotes the original (frozen) image encoder. The inner maximization seeks adversarial perturbations that distort the original embeddings, while the outer minimization encourages the model to preserve them under such perturbations.

\subsection{Analysis of Adversarial Attack Strategies in Fine-Tuning}
\label{subsec:analysis}

\begin{figure*}[t]
  \centering
  \includegraphics[width=0.90\linewidth]{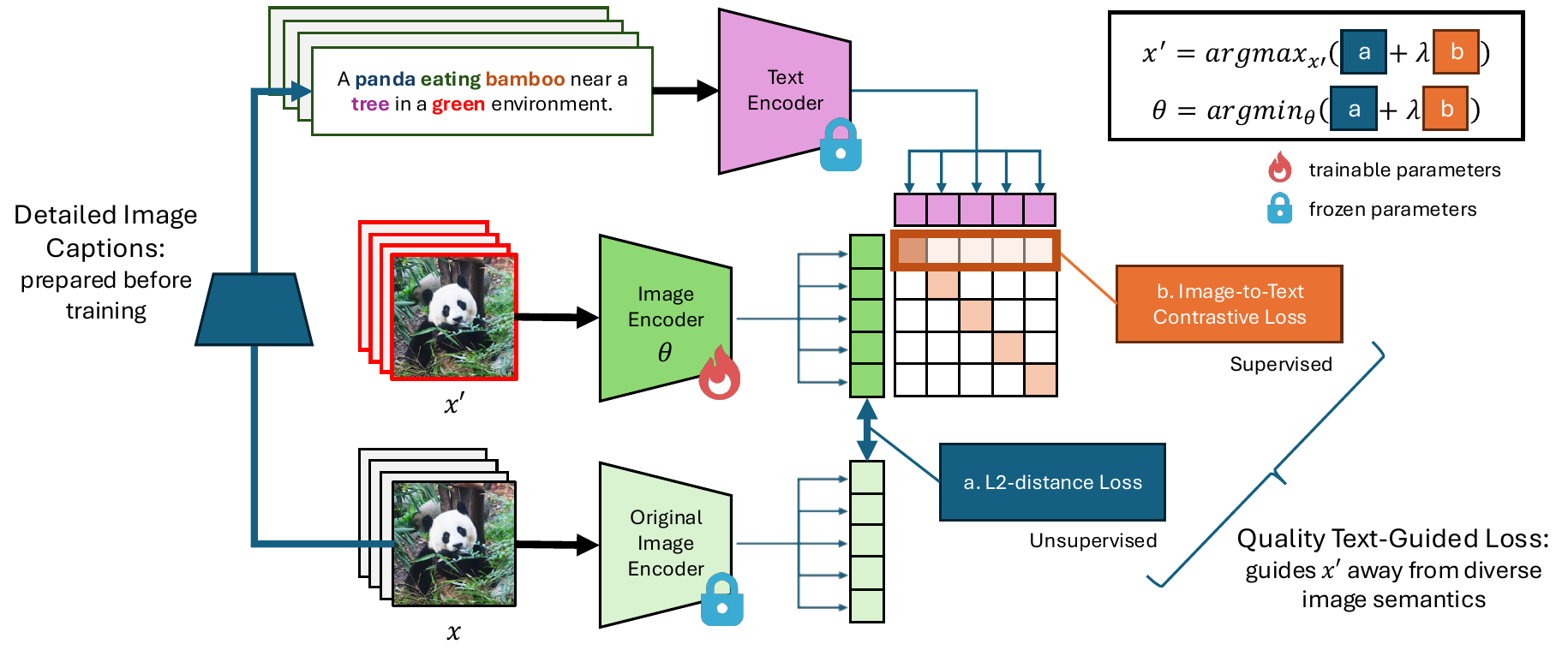}
  \caption{Our proposed method, Quality Text-guided Adversarial Fine-Tuning (QT-AFT), leverages rich captions instead of class labels to guide adversarial examples $x'$ away from diverse image semantics.
  The captions are pre-generated before training.
  By combining rich linguistic supervision with an unsupervised objective, we maximize separation from diverse semantics, enhancing robustness across diverse downstream zero-shot tasks.
  }
  \label{fig:qt-aft-method}
\end{figure*}

\begin{table}[t]
    \centering
    \caption{
    Cosine similarity between adversarial images and texts. TeCoA primarily minimizes similarity with class labels, while FARE minimizes similarity with the original image. In contrast, our method minimizes similarity with captions, promoting divergence from the diverse semantics.
    }
    \label{tab:analysis-abl}
    \resizebox{\columnwidth}{!}{
    \begin{tabular}{ll|r|r:r}
    \toprule
      & {(Cosine Similarity)}  & \multicolumn{1}{c}{{Image}} & \multicolumn{2}{c}{{Text}}  \\ \cmidrule(lr){3-3} \cmidrule(lr){4-5}
     & {Image} & \multicolumn{1}{c}{Clean} & \multicolumn{1}{c}{Label} & \multicolumn{1}{c}{\textbf{Caption}} \\
\midrule
\rowcolorbase & Clean & 1.000 \plusbar{1.000}{1.000} & 0.285 \plusbarB{0.285}{0.285} & 0.313 \plusbarC{0.313}{0.313} \\ \hline
     \multirow{5}{*}{\rotatebox{90}{Adv.}} & TeCoA ($\mathrm{Sup}_{\mathrm{label}}$) & 0.597 \plusbar{1.000}{0.597}  & \underline{0.120} \plusbarB{0.285}{0.120}  & 0.203 \plusbarC{0.313}{0.203}  \\
     & FARE (Unsup) &  \textbf{0.271} \plusbar{1.000}{0.271}  & 0.170 \plusbarB{0.285}{0.170}  & 0.157 \plusbarC{0.313}{0.157} \\ \cdashline{2-5}
     & ($\mathrm{Sup}_{\mathrm{caps}}$) & 0.576 \plusbar{1.000}{0.576} & 0.199 \plusbarB{0.285}{0.199} & \underline{0.100} \plusbarC{0.313}{0.100} \\
     & (Unsup + $\mathrm{Sup}_{\mathrm{label}}$) & 0.404 \plusbar{1.000}{0.404}  & \textbf{0.099} \plusbarB{0.285}{0.099}  & 0.155 \plusbarC{0.313}{0.155}  \\
     & \textbf{Ours (Unsup + $\mathrm{Sup}_{\mathrm{caps}}$)} & \underline{0.370} \plusbar{1.000}{0.370} & 0.171  \plusbarB{0.285}{0.171} & \textbf{0.091} \plusbarC{0.313}{0.091}  \\
     \bottomrule
    \end{tabular}
    }
    
\end{table}

How to generate AEs during AT plays a crucial role in achieving robustness, as these examples serve as data augmentation and directly influence the model's ability to resist perturbations.
In this work, we point out that both supervised AT based on class labels and unsupervised AT methods are suboptimal for achieving zero-shot adversarial robustness.

To investigate this, Tab.~\ref{tab:analysis-abl} analyzes how AEs deviate from textual representations. Specifically, we measure the cosine similarity between AEs and (i) the original image, (ii) the class label's text template (``a photo of \{\textit{class}\}''), and (iii) caption texts.
We conduct this analysis on ImageNet, using 10k randomly sampled images. Captions are synthetically generated using InternVL-2.5-8B~\cite{chen2024internvl}, and similarities are computed in CLIP's embedding space.
We compare the following AEs, each crafted using PGD with a different objective:
\begin{itemize}
    \item \textbf{TeCoA ($\mathrm{Sup}_{\mathrm{label}}$)}: A supervised attack that maximizes the cross-entropy loss between images and class label's templates (Eq.~\ref{eq:tecoa}).
    \item \textbf{FARE (Unsup)}: An unsupervised attack that maximizes the distance from the original images (Eq.~\ref{eq:fare}).
    \item $\mathrm{Sup}_{\mathrm{caps}}$: A supervised attack based on image captions, maximizing image-to-text CLIP loss  (Eq.~\ref{eq:clip_loss}).
    \item Unsup + $\mathrm{Sup}_{\mathrm{label}}$: A combinations of Unsup and $\mathrm{Sup}_{\mathrm{label}}$.
    \item \textbf{Ours  (Unsup + $\mathrm{Sup}_{\mathrm{caps}}$)}: A combination of the unsupervised objective (Eq.~\ref{eq:fare}) and a supervised objective of the CLIP loss between images and their \textbf{captions} (Eq.~\ref{eq:clip_loss}).
\end{itemize}

Tab.~\ref{tab:analysis-abl} demonstrates that AEs from \textbf{TeCoA} primarily reduce similarity to class labels, with minimal change relative to the original image or caption. This suggests TeCoA overfits to class templates, neglecting other semantics in the image, leading to suboptimal zero-shot robustness.
\textbf{FARE} reduces similarity not only to the original image but also to captions and class templates, demonstrating improved generalization beyond class labels. However, we argue that FARE overly focuses on diverging from the image representation, without fully disrupting the rich semantics present in the images.
In contrast, \textbf{our method} explicitly guides AEs to diverge from various semantic information present in images using \textit{captions}, while also leveraging the generalization benefits of the unsupervised objective. This dual-objective design aims to generate semantically challenging AEs that improve zero-shot robustness across varied downstream tasks.

We observe that, instead of using only the $\mathrm{Sup}_{\mathrm{caps}}$ objective, additionally incorporating the unsupervised objective provides better guidance for the adversarial direction, helping to minimize similarity with both class labels and captions.
Moreover, simply combining the unsupervised objective with $\mathrm{Sup}_{\mathrm{label}}$ fails to produce strong deviation from captions, highlighting the importance of directly leveraging caption information. 

\begin{table*}[t]
    \centering
    \caption{\textbf{Clean accuracy and robust accuracy against AutoAttack ($\epsilon=4/255$)} of CLIP, trained on ImageNet. Our method achieves state-of-the-art robustness and clean accuracy across a wide range of zero-shot datasets.}
     \label{tab:zeroshot}
    \resizebox{\textwidth}{!}{
    \begin{tabular}{cc|c|cccccccccccccccc|c}
    \toprule
    & & & \multicolumn{15}{c}{Zero-shot datasets} & \\ \cmidrule(lr){4-19}
      & Method & \rotvtwo{ImageNet} & \rotvtwo{ImageNet-S} & \rotvtwo{ImageNet-R} & \rotvtwo{CIFAR-10} & \rotvtwo{CIFAR-100} & \rotvtwo{STL-10} & \rotvtwo{Caltech101} & \rotvtwo{Caltech256} & \rotvtwo{OxfordPets} & \rotvtwo{Flowers102} & \rotvtwo{FGVC} & \rotvtwo{StanfordCars} & \rotvtwo{SUN397} & \rotvtwo{Food101} & \rotvtwo{EuroSAT} & \rotvtwo{DTD} & \rotvtwo{PCAM} & \rotvtwo{Avg. Zero-shot} \\
    \midrule
    \multirow{5}{*}{\rotatebox{90}{Clean}} & PMG-AFT & 55.6 & 31.7 & 50.9 & \underline{76.6} & 45.9 & 92.5 & 77.7 & 67.5 & 67.1 & 9.9 & 2.9 & 8.6 & 34.9 & 27.9 & \textbf{23.5} & 24.8 & 48.0 & 43.1 \\
& TGA-ZSR & \textbf{83.4} & \textbf{42.3} & 52.4 & \textbf{88.7} & \textbf{58.7} & \textbf{96.6} & 15.1 & \textbf{83.9} & 67.3 & \textbf{48.7} & 8.5 & 36.7 & \textbf{64.8} & \textbf{77.0} & 0.2 & \textbf{29.6} & 47.5 & \underline{51.1} \\
& TeCoA & \underline{63.3} & 31.8 & 51.9 & 75.2 & 39.1 & 91.7 & 74.7 & 66.4 & 71.8 & 19.5 & 6.9 & 12.6 & 34.7 & 20.9 & 17.0 & 21.4 & \textbf{57.9} & 43.3 \\
& FARE & 50.6 & 35.6 & \textbf{57.0} & 64.5 & 47.3 & 91.8 & \underline{80.5} & \underline{74.4} & \underline{76.4} & \underline{39.1} & \textbf{13.5} & \underline{39.5} & 43.8 & \underline{44.3} & \underline{21.9} & 27.0 & 48.0 & 50.3 \\
& \textbf{(ours) QT-AFT} & 51.9 & \underline{38.5} & \underline{56.9} & 70.9 & \underline{48.6} & \underline{95.8} & \textbf{81.9} & 73.4 & \textbf{80.7} & 30.6 & \underline{12.5} & \textbf{40.1} & \underline{51.7} & 44.2 & 19.2 & \underline{29.2} & \underline{51.1} & \textbf{51.6} \\
  \hline
        
    \multirow{5}{*}{\rotatebox{90}{Adv.}} & PMG-AFT & 30.1 & \underline{14.8} & 24.9 & \textbf{36.7} & \underline{17.7} & \textbf{70.3} & \underline{55.8} & 35.8 & \textbf{39.6} & 3.0 & 0.3 & 1.1 & 10.1 & 5.6 & 3.1 & 10.4 & \underline{47.7} & \underline{23.6} \\
& TGA-ZSR & \underline{31.1} & 3.2 & 10.9 & 24.5 & 8.5 & 54.4 & 10.3 & 30.0 & 13.5 & 2.8 & 0.0 & 0.1 & 0.5 & 5.0 & 0.0 & 0.4 & 0.0 & 10.3 \\
& TeCoA & \textbf{32.8} & 14.3 & \textbf{25.2} & 32.2 & 16.8 & 68.6 & 49.0 & \underline{36.3} & \underline{39.2} & 5.8 & 1.2 & 2.6 & \underline{10.5} & 6.0 & \underline{9.8} & 10.2 & 20.5 & 21.8 \\
& FARE & 20.0 & 14.0 & 20.9 & 30.7 & 15.2 & 62.6 & 53.0 & 35.8 & 30.4 & \underline{8.6} & \underline{1.8} & \underline{2.8} & 9.8 & \underline{7.4} & 3.6 & \underline{13.2} & \textbf{48.0} & 22.4 \\
& \textbf{(ours) QT-AFT} & 19.6 & \textbf{17.6} & 25.2 & \underline{33.2} & \textbf{20.9} & \underline{69.0} & \textbf{58.9} & \textbf{40.6} & 36.5 & \textbf{9.7} & \textbf{2.3} & \textbf{5.8} & \textbf{14.2} & \textbf{7.7} & \textbf{12.6} & \textbf{14.6} & 44.1 & \textbf{25.8} \\

    \bottomrule
    \end{tabular}
    }

\end{table*}

\subsection{Quality Text-guided Adversarial Fine-Tuning (QT-AFT)}
\label{subsec:qt-aft}

Based on the findings in the previous section, we propose to leverage high-quality image captions during adversarial fine-tuning, introducing \textbf{Quality Text-guided Adversarial Fine-Tuning (QT-AFT)} (Fig.~\ref{fig:qt-aft-method}). Our method consists of two steps: (1) caption preparation and (2) adversarial fine-tuning guided by quality captions.

\paragraph{Step 1.} Captions can be sourced in various ways, including human annotations, image-to-text models, or web-scraped descriptions. In this work, for reproducibility and controllability, we generate synthetic captions using a VL multimodal model with the prompt: ``Describe the image in detail within 50 words.'' We constrain the captions to approximately 50 words to ensure compatibility with the CLIP text encoder, which has a limited token capacity of 77 tokens ($\sim$ 50 words).

\paragraph{Step 2.} The objective function for QT-AFT is defined as follows: 
\begin{align}
    \mathcal{L}_{\text{QT-AFT}}(x',t) = \sum_{i=1}^N \Bigg[ 
    & \left\| f_\theta(x_i') - f_{{\theta}_{\text{orig}}}(x_i) \right\|^2_2 \nonumber \\
    & - \lambda \cdot \log \frac{
        \exp\left(\cos(f_{\theta}(x'_i), f_{\phi}(t_i)) / \tau\right)
    }{
        \sum_{j=1}^N \exp\left(\cos(f_{\theta}(x'_i), f_{\phi}(t_j)) / \tau\right)
    } \Bigg],
    \label{eq:ours}
\end{align}
\begin{align}
    \theta = \arg\min_{\theta} \mathbb{E}_{(x,t) \sim \mathcal{D}} 
    \left[ \max_{x' \in B(x,\epsilon)} \mathcal{L}_{\text{QT-AFT}}(x', t) \right].
    \label{eq:ours-theta}
\end{align}
Here, $t_i$ denotes the caption generated from image $x_i$, and $\lambda$ is the hyperparameter. In Eq.~\ref{eq:ours}, the first term represents the unsupervised objective, while the second term applies a VL contrastive loss using captions, and their effective combination encourages deviation from the caption representations.

\section{Experiments}

\subsection{Experimental settings}

\paragraph{\textbf{Model and Datasets.}}
We fine-tune CLIP-ViT-B/16~\cite{radford2021learning} on ImageNet~\cite{deng2009imagenet} and evaluate its zero-shot performance on a wide range of image classification datasets.
Additionally, we fine-tune CLIP-ViT-L/14 for the ablation study.
To generate captions for ImageNet, we use InternVL-2.5-8B~\cite{chen2024internvl}, a state-of-the-art VL multimodal model (captions will be released publicly).
For zero-shot performance, we evaluate on 16 datasets across six categories; ImageNet style variants such as ImageNet-S~\cite{wang2019learning} (sketch style) and ImageNet-R~\cite{hendrycks2021many} (diverse styles); general object recognition including CIFAR10~\cite{krizhevsky2009learning_cifar10}, CIFAR100~\cite{krizhevsky2009learning_cifar10},
STL10~\cite{coates2011analysis}, Caltech101~\cite{fei2006one}, and Caltech256~\cite{griffin2007caltech}; fine-grained recognition such as OxfordPets~\cite{parkhi2012cats}, Flowers102~\cite{nilsback2008automated}, FGVCAircraft~\cite{maji2013fine_fgvc_aircraft}, and StanfordCars~\cite{krause20133d}; scene recognition represented by SUN397~\cite{xiao2010sun};
domain-specific tasks such as Food101~\cite{bossard2014food}, EuroSAT~\cite{helber2019eurosat}, and DTD~\cite{cimpoi2014describing}; medical imaging,
PCAM~\cite{bejnordi2017diagnostic}.
All images from the evaluated datasets are resized to a resolution of $3 \times 224 \times 224$.


\paragraph{\textbf{Implementation details.}}
For adversarial fine-tuning, we train for two epochs with an initial learning rate of 1e-5, decayed using cosine scheduling. We use the AdamW optimizer with a weight decay of 1e-4 and a batch size of 128. AEs are generated using 10-step PGD with a perturbation size of $\epsilon = 4/255$ under the $\ell_{\infty}$-norm and a step size of $1/255$. The hyperparameter $\lambda$ in Eq.~\ref{eq:ours} is set to 10. 
For reliable evaluation, we evaluate against AutoAttack~\cite{croce2020reliable}, using the same perturbation size of $\epsilon = 4/255$. Due to its high computational cost, we perform the evaluation on 1,000 randomly selected samples for each dataset, following \citet{schlarmann2024robust}.
We present the evaluation for full samples using 10-step PGD in Appendix~\ref{sec:pgd-10}.

\begin{table*}[t]
    \centering
    \caption{\textbf{Caption Quality Analysis: Label vs. Caption.} Clean accuracy and robust accuracy against AutoAttack ($\epsilon=4/255$) of CLIP trained on ImageNet with QT-AFT, using either class labels or captions as supervision.
    Using captions outperforms using class labels, highlighting the benefit of referencing richer visual features through text. }
     \label{tab:abl-label-vs-caption}
    \resizebox{\textwidth}{!}{
    \begin{tabular}{cc|c|cccccccccccccccc|c}
    \toprule
    & & & \multicolumn{16}{c}{Zero-shot datasets} & \\ \cmidrule(lr){4-19}
      & Method & \rotvtwo{ImageNet} & \rotvtwo{ImageNet-S} & \rotvtwo{ImageNet-R} & \rotvtwo{CIFAR-10} & \rotvtwo{CIFAR-100} & \rotvtwo{STL-10} & \rotvtwo{Caltech101} & \rotvtwo{Caltech256} & \rotvtwo{OxfordPets} & \rotvtwo{Flowers102} & \rotvtwo{FGVC} & \rotvtwo{StanfordCars} & \rotvtwo{SUN397} & \rotvtwo{Food101} & \rotvtwo{EuroSAT} & \rotvtwo{DTD} & \rotvtwo{PCAM} & \rotvtwo{Avg. Zero-shot} \\
    \midrule
    \multirow{3}{*}{\rotatebox{90}{Clean}} & FARE & 50.6 & 35.6 & \textbf{57.0} & 64.5 & 47.3 & \underline{91.8} & 80.5 & \textbf{74.4} & \underline{76.4} & \textbf{39.1} & \textbf{13.5} & \underline{39.5} & \underline{43.8} & \textbf{44.3} & \underline{21.9} & 27.0 & 48.0 & \underline{50.3} \\ \cdashline{2-20}
& \textbf{(ours) QT-AFT w/ label} & \textbf{58.0} & \textbf{38.8} & 57.0 & \textbf{75.0} & \textbf{53.1} & 90.4 & \underline{81.3} & \underline{73.5} & 71.3 & \underline{35.9} & 11.5 & 27.4 & 42.9 & 33.0 & \textbf{22.1} & \textbf{30.2} & \textbf{51.1} & 49.7 \\
& \textbf{(ours) QT-AFT w/ caps (default)} & \underline{51.9} & \underline{38.5} & 56.9 & \underline{70.9} & \underline{48.6} & \textbf{95.8} & \textbf{81.9} & 73.4 & \textbf{80.7} & 30.6 & \underline{12.5} & \textbf{40.1} & \textbf{51.7} & \underline{44.2} & 19.2 & \underline{29.2} & 51.1 & \textbf{51.6} \\

 \hline

    \multirow{3}{*}{\rotatebox{90}{Adv.}} & FARE & \underline{20.0} & 14.0 & 20.9 & 30.7 & 15.2 & 62.6 & \underline{53.0} & 35.8 & 30.4 & \underline{8.6} & 1.8 & 2.8 & 9.8 & \underline{7.4} & 3.6 & \underline{13.2} & \textbf{48.0} & 22.4 \\  \cdashline{2-20}
& \textbf{(ours) QT-AFT w/ label} & \textbf{26.9} & \underline{16.9} & \underline{22.8} & \underline{32.4} & \textbf{20.9} & \underline{62.7} & 50.8 & \underline{37.3} & \textbf{36.5} & 6.6 & \underline{2.1} & \underline{4.6} & \underline{11.7} & 6.2 & \underline{12.2} & 9.4 & 40.0 & \underline{23.3} \\
& \textbf{(ours) QT-AFT w/ caps (default)} & 19.6 & \textbf{17.6} & \textbf{25.2} & \textbf{33.2} & 20.9 & \textbf{69.0} & \textbf{58.9} & \textbf{40.6} & 36.5 & \textbf{9.7} & \textbf{2.3} & \textbf{5.8} & \textbf{14.2} & \textbf{7.7} & \textbf{12.6} & \textbf{14.6} & \underline{44.1} & \textbf{25.8} \\

    \bottomrule
    \end{tabular}
    }

\end{table*}

\begin{table*}[t]
    \centering
    \caption{\textbf{Caption Quality Analysis: Image Captioning Model.} Clean accuracy and AutoAttack robustness ($\epsilon=4/255$) of CLIP trained on ImageNet with QT-AFT, using captions generated by different VL models. Using a smaller VL model to generate captions also achieves state-of-the-art performance; however, leveraging a stronger captioning model achives the best results.}
     \label{tab:abl-captioning-model}
    \resizebox{\textwidth}{!}{
    \begin{tabular}{cc|c|cccccccccccccccc|c}
    \toprule
    & & & \multicolumn{16}{c}{Zero-shot datasets} & \\ \cmidrule(lr){4-19}
      & Method & \rotvtwo{ImageNet} & \rotvtwo{ImageNet-S} & \rotvtwo{ImageNet-R} & \rotvtwo{CIFAR-10} & \rotvtwo{CIFAR-100} & \rotvtwo{STL-10} & \rotvtwo{Caltech101} & \rotvtwo{Caltech256} & \rotvtwo{OxfordPets} & \rotvtwo{Flowers102} & \rotvtwo{FGVC} & \rotvtwo{StanfordCars} & \rotvtwo{SUN397} & \rotvtwo{Food101} & \rotvtwo{EuroSAT} & \rotvtwo{DTD} & \rotvtwo{PCAM} & \rotvtwo{Avg. Zero-shot} \\
    \midrule
    \multirow{2}{*}{\rotatebox{90}{Clean}} & \textbf{QT-AFT w/ Mini-IntVL-1.5-2B} & \textbf{52.1} & \underline{36.2} & \textbf{57.8} & \textbf{73.8} & \textbf{49.0} & \underline{93.8} & \textbf{82.8} & \textbf{75.6} & \underline{75.0} & \textbf{33.8} & \underline{11.5} & \underline{38.0} & \underline{48.3} & \textbf{44.8} & \textbf{19.6} & \underline{21.9} & \textbf{52.0} & \underline{50.9} \\
& \textbf{QT-AFT w/ IntVL-2.5-8B (default)} & \underline{51.9} & \textbf{38.5} & \underline{56.9} & \underline{70.9} & \underline{48.6} & \textbf{95.8} & \underline{81.9} & \underline{73.4} & \textbf{80.7} & \underline{30.6} & \textbf{12.5} & \textbf{40.1} & \textbf{51.7} & \underline{44.2} & \underline{19.2} & \textbf{29.2} & \underline{51.1} & \textbf{51.6} \\
 \hline

    \multirow{2}{*}{\rotatebox{90}{Adv.}} 
& \textbf{QT-AFT w/ Mini-IntVL-1.5-2B} & \textbf{20.5} & \underline{15.2} & \underline{22.4} & \underline{32.4} & \underline{19.1} & \underline{67.1} & \underline{54.5} & \textbf{42.0} & \textbf{36.5} & \textbf{10.9} & \underline{1.0} & \underline{5.3} & \underline{13.0} & \textbf{8.5} & \underline{12.4} & \underline{6.2} & \textbf{49.3} & \underline{24.7} \\
& \textbf{QT-AFT w/ IntVL-2.5-8B (default)} & \underline{19.6} & \textbf{17.6} & \textbf{25.2} & \textbf{33.2} & \textbf{20.9} & \textbf{69.0} & \textbf{58.9} & \underline{40.6} & \textbf{36.5} & \underline{9.7} & \textbf{2.3} & \textbf{5.8} & \textbf{14.2} & \underline{7.7} & \textbf{12.6} & \textbf{14.6} & \underline{44.1} & \textbf{25.8} \\
    \bottomrule
    \end{tabular}
    }

\end{table*}

\subsection{Results}

\textbf{QT-AFT Achieves State-of-the-Art Performance.}
Tab.~\ref{tab:zeroshot} compares clean and robust accuracy across 16 zero-shot datasets. Our proposed QT-AFT achieves state-of-the-art zero-shot robustness on 12 out of 16 datasets, with an average improvement of more than 2\%. Notably, QT-AFT also retains high clean accuracy, achieving state-of-the-art clean accuracy on average.

\textbf{Class label-based baselines overfit to training distribution, while QT-AFT does not.}
The existing supervised baselines, TeCoA, PMG-AFT, and TGA-ZSR, which uses class labels, achieve strong robustness on the training dataset (ImageNet), but their performance on zero-shot datasets is limited.
For example, while TeCoA and PMG-AFT achieve over 30\% robustness on ImageNet, surpassing FARE and QT-AFT by 10\%, they show poor zero-shot clean accuracy of around 43\%, which is 7\% lower than FARE and QT-AFT.
These results suggest that while leveraging class labels during adversarial fine-tuning enhances robustness on the training distribution, it may limit zero-shot performance due to overfitting.
On the other hand, QT-AFT effectively addresses this limitation by incorporating image captions as supervision during adversarial fine-tuning, avoiding overfitting to class labels.

\textbf{QT-AFT Outperforms FARE in Robustness.}
FARE avoids overfitting and maintains high clean accuracy, achieving 7\% higher accuracy compared to supervised AT baselines. However, QT-AFT further addresses FARE's limitation—its lack of semantic awareness during AT—by leveraging rich linguistic guidance, leading to enhanced robustness. On average, QT-AFT improves zero-shot robustness by over 3\% and clean accuracy by 1\%. This demonstrates the effectiveness of our approach in generating AEs that deviate from diverse image semantics, enhancing robustness across a variety of zero-shot tasks.

\section{Analysis: Impact of Caption Quality}

In this section, to better understand our proposed method, we conduct a comprehensive study on the impact of caption quality. Specifically, we compare the performance of QT-AFT using different types of captions.

\subsection{Label vs. Caption: Using Captions Outperforms Class Labels} 

In our proposed method, QT-AFT, we use image captions as supervision during AT. To evaluate the impact of caption quality, we replace the caption-guided supervised objective with a class label-guided objective, following the approach used in TeCoA. This corresponds to combining the unsupervised FARE loss with the class-label-based TeCoA loss.
We then compare the effectiveness of using class labels versus captions for supervision.

As shown in Tab.~\ref{tab:abl-label-vs-caption}, simply adding a class-label-guided objective to the unsupervised loss already improves the performance of FARE. However, using captions leads to significantly better results compared to using labels. This highlights the value of captions as explicit semantic guidance for generating AEs during AT to enhance zero-shot robustness.

\begin{figure}[t]
  \centering
  \begin{minipage}[t]{0.45\columnwidth}
    \centering
    \includegraphics[width=\linewidth]{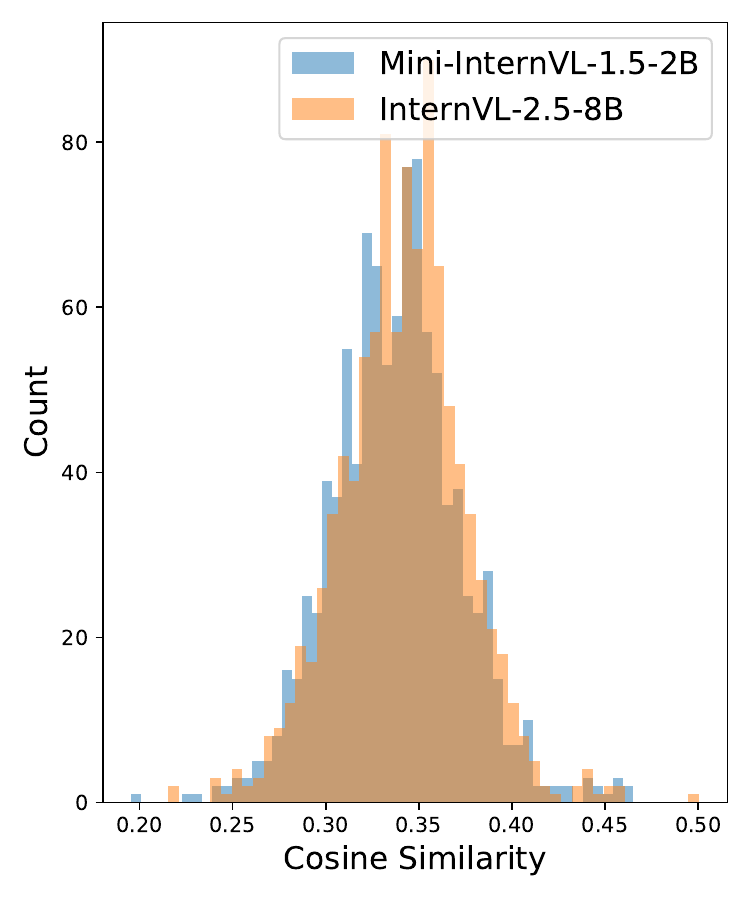}
    \subcaption{Image-caption similarity}
    \label{fig:analysis-internvl-cos-sim}
  \end{minipage}
  \hfill
  \begin{minipage}[t]{0.45\columnwidth}
    \centering
    \includegraphics[width=\linewidth]{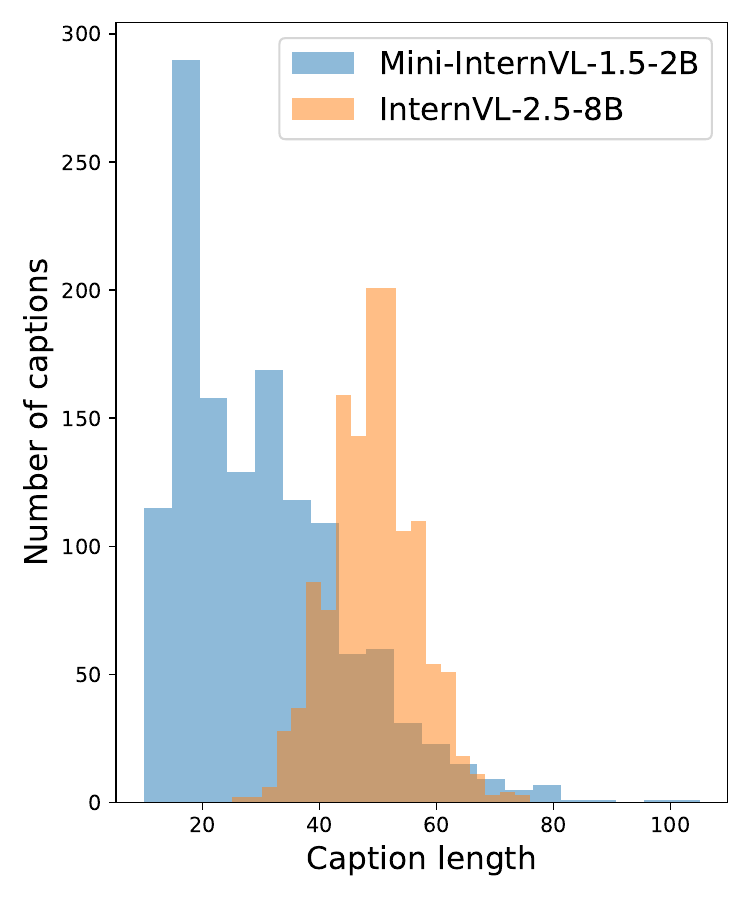}
    \subcaption{Caption length}
    \label{fig:analysis-internvl-cap-length}
  \end{minipage}
  \caption{Analysis on the image caption quality generated by different VL models.}
  \vspace{-12pt}
\end{figure}

\begin{table*}[ht]
    \centering
    \caption{\textbf{Caption Quality Analysis: Word Class.} Clean accuracy and AutoAttack robustness ($\epsilon=4/255$) of CLIP trained on ImageNet, with QT-AFT using different caption modifications.
    Each setting alters the original full caption (default).
    Cells with green backgrounds indicate improved accuracy compared to the original caption, while red backgrounds indicate degradation.
    }
     \label{tab:abl-text-type}
    \resizebox{\textwidth}{!}{
    \begin{tabular}{cr|c|cccccccccccccccc|c}
    \toprule
    & & & \multicolumn{16}{c}{Zero-shot datasets} & \\ \cmidrule(lr){4-19}
      & Method & \rotvtwo{ImageNet} & \rotvtwo{ImageNet-S} & \rotvtwo{ImageNet-R} & \rotvtwo{CIFAR-10} & \rotvtwo{CIFAR-100} & \rotvtwo{STL-10} & \rotvtwo{Caltech101} & \rotvtwo{Caltech256} & \rotvtwo{OxfordPets} & \rotvtwo{Flowers102} & \rotvtwo{FGVC} & \rotvtwo{StanfordCars} & \rotvtwo{SUN397} & \rotvtwo{Food101} & \rotvtwo{EuroSAT} & \rotvtwo{DTD} & \rotvtwo{PCAM} & \rotvtwo{Avg. Zero-shot} \\
    \midrule
    
    \rowcolorbase \multirow{7}{*}{\rotatebox{90}{Clean}} & \textbf{(ours) QT-AFT} & 51.9 & \underline{38.5} & 56.9 & 70.9 & 48.6 & \textbf{95.8} & 81.9 & 73.4 & \underline{80.7} & 30.6 & 12.5 & 40.1 & \textbf{51.7} & \textbf{44.2} & 19.2 & 29.2 & 51.1 & \underline{51.6} \\
& Nouns-only & \cellcolor{green!20} \textbf{55.4} & \cellcolor{green!20} \textbf{40.2} & \cellcolor{green!20} \underline{59.7} & \cellcolor{green!20} 71.7 & \cellcolor{green!20} 48.6 & \cellcolor{red!20} 94.5 & \cellcolor{green!20} 83.0 & \cellcolor{green!20} \underline{75.9} & \cellcolor{red!20} 78.1 & \cellcolor{green!20} 31.9 & \cellcolor{red!20} 10.2 & \cellcolor{red!20} 35.1 & \cellcolor{red!20} \underline{47.8} & \cellcolor{red!20} 40.2 & \cellcolor{green!20} \underline{20.6} & \cellcolor{green!20} \underline{29.2} & \cellcolor{green!20} 51.4 & \cellcolor{red!20} 51.1 \\
& No adj./adv. & \cellcolor{red!20} 50.5 & \cellcolor{red!20} 38.2 & \cellcolor{green!20} \textbf{61.2} & \cellcolor{green!20} \textbf{72.1} & \cellcolor{green!20} 49.2 & \cellcolor{red!20} 93.0 & \cellcolor{green!20} 83.3 & \cellcolor{red!20} 72.6 & \cellcolor{green!20} \textbf{81.2} & \cellcolor{red!20} 29.4 & \cellcolor{green!20} \textbf{14.8} & \cellcolor{green!20} \textbf{42.1} & \cellcolor{red!20} 44.5 & \cellcolor{red!20} 43.1 & \cellcolor{red!20} 18.2 & \cellcolor{red!20} 24.0 & \cellcolor{green!20} 51.1 & \cellcolor{red!20} 51.1 \\
& No nouns & \cellcolor{green!20} \underline{52.6} & \cellcolor{red!20} 34.6 & \cellcolor{red!20} 54.8 & \cellcolor{red!20} 68.5 & \cellcolor{red!20} 47.7 & \cellcolor{red!20} 92.2 & \cellcolor{green!20} \underline{84.4} & \cellcolor{green!20} 74.1 & \cellcolor{red!20} 75.0 & \cellcolor{green!20} \textbf{33.1} & \cellcolor{red!20} 11.9 & \cellcolor{red!20} 35.8 & \cellcolor{red!20} 47.0 & \cellcolor{red!20} 41.3 & \cellcolor{green!20} 19.4 & \cellcolor{red!20} 27.1 & \cellcolor{green!20} \underline{52.2} & \cellcolor{red!20} 49.9 \\
& No function words & \cellcolor{red!20} 51.6 & \cellcolor{red!20} 34.1 & \cellcolor{green!20} 59.5 & \cellcolor{red!20} 70.5 & \cellcolor{red!20} 47.7 & \cellcolor{red!20} \underline{94.8} & \cellcolor{green!20} \textbf{85.5} & \cellcolor{green!20} \textbf{76.3} & \cellcolor{red!20} 75.0 & \cellcolor{green!20} \underline{32.8} & \cellcolor{green!20} 12.5 & \cellcolor{red!20} 37.7 & \cellcolor{red!20} 47.1 & \cellcolor{red!20} \underline{43.8} & \cellcolor{red!20} 18.6 & \cellcolor{red!20} 26.0 & \cellcolor{red!20} 49.4 & \cellcolor{red!20} 50.7 \\
& Shuffle words & \cellcolor{green!20} 52.2 & \cellcolor{red!20} 35.7 & \cellcolor{green!20} 57.8 & \cellcolor{green!20} 71.5 & \cellcolor{green!20} \textbf{54.1} & \cellcolor{red!20} 91.2 & \cellcolor{green!20} 82.8 & \cellcolor{green!20} 74.4 & \cellcolor{red!20} 74.5 & \cellcolor{red!20} 26.6 & \cellcolor{green!20} \underline{13.6} & \cellcolor{green!20} \underline{40.6} & \cellcolor{red!20} 45.8 & \cellcolor{red!20} 39.6 & \cellcolor{green!20} 19.6 & \cellcolor{red!20} 20.8 & \cellcolor{red!20} 50.4 & \cellcolor{red!20} 49.9 \\

 \hline

    \rowcolorbase \multirow{7}{*}{\rotatebox{90}{Adv.}} & \textbf{(ours) QT-AFT} & 19.6 & 17.6 & 25.2 & 33.2 & \underline{20.9} & 69.0 & \textbf{58.9} & 40.6 & \underline{36.5} & \underline{9.7} & 2.3 & \textbf{5.8} & \textbf{14.2} & \underline{7.7} & 12.6 & 14.6 & 44.1 & \textbf{25.8} \\
& Nouns-only & \cellcolor{green!20} 20.9 & \cellcolor{green!20} \textbf{18.9} & \cellcolor{green!20} \underline{26.4} & \cellcolor{red!20} 32.4 & \cellcolor{red!20} 20.3 & \cellcolor{green!20} \textbf{70.2} & \cellcolor{red!20} 54.0 & \cellcolor{red!20} 40.5 & \cellcolor{red!20} 32.8 & \cellcolor{red!20} 8.8 & \cellcolor{red!20} 1.1 & \cellcolor{red!20} 5.5 & \cellcolor{red!20} 13.7 & \cellcolor{red!20} 7.2 & \cellcolor{green!20} \textbf{14.7} & \cellcolor{green!20} 14.6 & \cellcolor{green!20} \textbf{48.4} & \cellcolor{red!20} 25.6 \\
& No adj./adv. & \cellcolor{green!20} 19.9 & \cellcolor{green!20} \underline{18.2} & \cellcolor{green!20} 25.9 & \cellcolor{green!20} \underline{33.2} & \cellcolor{red!20} 16.8 & \cellcolor{red!20} 64.2 & \cellcolor{red!20} 57.1 & \cellcolor{red!20} 37.5 & \cellcolor{red!20} 28.1 & \cellcolor{red!20} 7.8 & \cellcolor{green!20} \underline{2.8} & \cellcolor{green!20} \underline{5.8} & \cellcolor{red!20} 13.4 & \cellcolor{red!20} 7.5 & \cellcolor{red!20} 11.5 & \cellcolor{red!20} 12.5 & \cellcolor{green!20} 46.7 & \cellcolor{red!20} 24.3 \\
& No nouns & \cellcolor{red!20} 19.0 & \cellcolor{red!20} 14.9 & \cellcolor{red!20} 23.0 & \cellcolor{red!20} 31.4 & \cellcolor{red!20} 18.2 & \cellcolor{green!20} \underline{70.0} & \cellcolor{red!20} 56.5 & \cellcolor{red!20} 38.8 & \cellcolor{red!20} 29.7 & \cellcolor{red!20} 5.9 & \cellcolor{red!20} 0.6 & \cellcolor{red!20} 4.8 & \cellcolor{red!20} 11.3 & \cellcolor{red!20} 5.8 & \cellcolor{red!20} 10.9 & \cellcolor{green!20} \textbf{20.8} & \cellcolor{red!20} 29.1 & \cellcolor{red!20} 23.2 \\
& No function words & \cellcolor{green!20} \textbf{21.6} & \cellcolor{red!20} 16.7 & \cellcolor{green!20} 26.2 & \cellcolor{green!20} \textbf{33.4} & \cellcolor{red!20} 19.3 & \cellcolor{red!20} 67.2 & \cellcolor{red!20} \underline{58.7} & \cellcolor{green!20} \textbf{43.7} & \cellcolor{green!20} \textbf{37.0} & \cellcolor{red!20} 7.2 & \cellcolor{green!20} \textbf{3.4} & \cellcolor{red!20} 3.1 & \cellcolor{red!20} 13.4 & \cellcolor{red!20} 6.7 & \cellcolor{red!20} 11.4 & \cellcolor{red!20} 12.5 & \cellcolor{red!20} 38.8 & \cellcolor{red!20} 24.9 \\
& Shuffle words & \cellcolor{green!20} \underline{21.1} & \cellcolor{red!20} 17.4 & \cellcolor{green!20} 25.5 & \cellcolor{red!20} 33.0 & \cellcolor{green!20} \textbf{21.9} & \cellcolor{red!20} 63.2 & \cellcolor{red!20} 56.7 & \cellcolor{green!20} \underline{40.6} & \cellcolor{red!20} 28.6 & \cellcolor{red!20} 7.2 & \cellcolor{red!20} 1.7 & \cellcolor{green!20} 5.8 & \cellcolor{red!20} 13.4 & \cellcolor{red!20} 6.8 & \cellcolor{green!20} \underline{13.9} & \cellcolor{red!20} 9.4 & \cellcolor{red!20} 38.3 & \cellcolor{red!20} 24.0 \\

    \bottomrule
    \end{tabular}
    }
     
\end{table*}

\begin{figure*}[ht]
\centering

\begin{minipage}{0.14\linewidth}
    \centering
    \includegraphics[width=\linewidth]{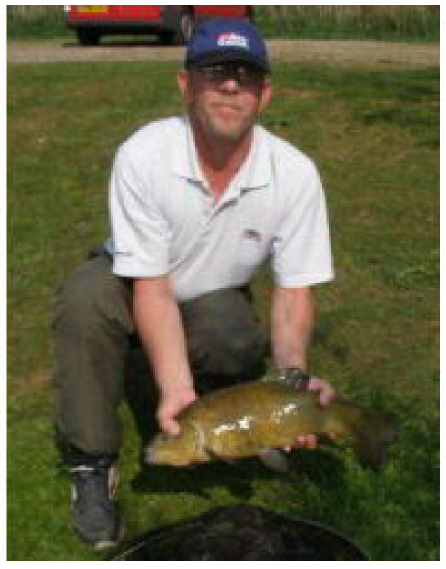}
\end{minipage} \hfill \hspace{-20pt}
\begin{minipage}{0.83\linewidth}
    \centering
    \small
    \begin{tcolorbox}[colframe=black, colback=white, coltitle=black, width=1.0\linewidth]
    \colorbox[rgb]{0.6,1.0,0.6}{A} \colorbox[rgb]{1.0,0.6,0.6}{man} \colorbox[rgb]{1.0,1.0,1.0}{kneels} \colorbox[rgb]{0.6,1.0,0.6}{on} \colorbox[rgb]{1.0,0.6,0.6}{grass} \colorbox[rgb]{1.0,1.0,1.0}{holding} \colorbox[rgb]{0.6,1.0,0.6}{a} \colorbox[rgb]{0.4,0.7,1.0}{large} \colorbox[rgb]{1.0,0.6,0.6}{fish} \colorbox[rgb]{0.6,1.0,0.6}{above} \colorbox[rgb]{0.4,0.7,1.0}{grassy} \colorbox[rgb]{1.0,0.6,0.6}{field.} \colorbox[rgb]{0.6,1.0,0.6}{He} \colorbox[rgb]{1.0,1.0,1.0}{wears} \colorbox[rgb]{1.0,0.6,0.6}{sunglasses} \colorbox[rgb]{0.6,1.0,0.6}{and} \colorbox[rgb]{1.0,0.6,0.6}{cap,} \colorbox[rgb]{1.0,1.0,1.0}{dressed} \colorbox[rgb]{0.6,1.0,0.6}{in} \colorbox[rgb]{0.4,0.7,1.0}{white} \colorbox[rgb]{1.0,0.6,0.6}{polo} \colorbox[rgb]{1.0,0.6,0.6}{shirt} \colorbox[rgb]{0.4,0.7,1.0}{camouflage} \colorbox[rgb]{1.0,0.6,0.6}{pants,} \colorbox[rgb]{1.0,1.0,1.0}{sporting} \colorbox[rgb]{0.4,0.7,1.0}{grey} \colorbox[rgb]{1.0,0.6,0.6}{sneakers.} \colorbox[rgb]{0.6,1.0,0.6}{In} \colorbox[rgb]{0.4,0.7,1.0}{front} \colorbox[rgb]{0.6,1.0,0.6}{of} \colorbox[rgb]{0.6,1.0,0.6}{him,} \colorbox[rgb]{1.0,0.6,0.6}{net} \colorbox[rgb]{0.6,1.0,0.6}{or} \colorbox[rgb]{1.0,0.6,0.6}{container} \colorbox[rgb]{1.0,1.0,1.0}{rests} \colorbox[rgb]{0.6,1.0,0.6}{the} \colorbox[rgb]{1.0,0.6,0.6}{ground,} \colorbox[rgb]{0.4,0.7,1.0}{likely} \colorbox[rgb]{1.0,1.0,1.0}{used} \colorbox[rgb]{0.6,1.0,0.6}{for} \colorbox[rgb]{1.0,0.6,0.6}{catch.} \colorbox[rgb]{0.6,1.0,0.6}{The} \colorbox[rgb]{0.4,0.7,1.0}{green} \colorbox[rgb]{1.0,0.6,0.6}{hues} \colorbox[rgb]{1.0,1.0,1.0}{contrast} \colorbox[rgb]{0.6,1.0,0.6}{with} \colorbox[rgb]{1.0,0.6,0.6}{man}\colorbox[rgb]{0.6,1.0,0.6}{’s} \colorbox[rgb]{0.4,0.7,1.0}{light} \colorbox[rgb]{1.0,0.6,0.6}{clothing} \colorbox[rgb]{0.4,0.7,1.0}{dark} \colorbox[rgb]{1.0,0.6,0.6}{fish.}

    \vspace{-10pt}
    \raggedleft
    (\colorbox[rgb]{1.0,0.6,0.6}{Red} = Nouns, 
    \colorbox[rgb]{0.4,0.7,1.0}{Blue} = Adjectives/Adverbs, 
    \colorbox[rgb]{0.6,1.0,0.6}{Green} = Function Words)
    \end{tcolorbox}
\end{minipage} 
\caption{An example caption along with its part-of-speech (POS) tags. In our word class analysis, for instance, the ``Nouns-only'' setting refers to removing all non-noun words and concatenating the remaining nouns.}
\label{fig:word-class-analysis}
\end{figure*}

\subsection{Image Captioning Model: The Benefits of Describing More Visual Features}

In our proposed method, we use InternVL-2.5-8B~\cite{chen2024internvl} to generate synthetic captions for ImageNet. To assess the impact of caption quality, we also experimented with Mini-InternVL-Chat-2B-V1-5~\cite{chen2024internvl}, a smaller variant with one-fourth the parameters and lower captioning performance.

Tab.~\ref{tab:abl-captioning-model} demonstrates that using the smaller VL model, Mini-InternVL-Chat-2B-V1-5, for QT-AFT still achieves state-of-the-art robustness and accuracy on the zero-shot datasets, further reinforcing the effectiveness of our approach.
However, it also shows that using the weaker captioning model, Mini-InternVL-Chat-2B-V1-5, leads to lower performance, reducing robustness on 11 out of 16 datasets. This suggests that caption quality plays a crucial role.
To better understand this, we analyze caption quality from two perspectives: (1) the cosine similarity between image and caption embeddings measured in CLIP’s embedding space, and (2) caption length.
Fig.~\ref{fig:analysis-internvl-cos-sim} shows the cosine similarity distributions between images and captions for both models, while Fig.~\ref{fig:analysis-internvl-cap-length} compares their caption lengths. Interestingly, the similarity distributions are nearly identical, indicating that CLIP-based image-text similarity is not the primary factor contributing to robustness gains. In contrast, we observe a clear difference in caption length: Mini-InternVL-Chat-2B-V1-5 tends to generate shorter captions, whereas InternVL-2.5-8B produces longer, more detailed descriptions—typically around 50 words, accurately following the prompt ``Describe the image in detail within 50 words.'' Despite using the same prompt, Mini-InternVL often lacks rich semantic content, possibly due to limited ability to follow the prompt—prioritizing brevity over detail—or difficulty in understanding image details.
These results suggest that mentioning more visual features contributes to improved visual robustness.
Please see the qualitative comparison of the generated captions in Appendix~\ref{sec:image-captioning-examples}.


\begin{table*}[ht]
    \centering
    \caption{\textbf{Ablation study: CLIP-ViT-L/14.} Clean accuracy and robust accuracy against AutoAttack ($\epsilon=4/255$). Our method achieves significantly higher clean accuracy while maintaining strong robustness, outperforming baselines on 8 out of 12 datasets.}
     \label{tab:abl-clip-l14}
    \resizebox{0.9\textwidth}{!}{
    \begin{tabular}{cc|c|cccccccccccc|c}
    \toprule
    & & & \multicolumn{11}{c}{Zero-shot datasets} & \\ \cmidrule(lr){4-15} 
      & Method & \rotvtwo{ImageNet} & \rotvtwo{CIFAR-10} & \rotvtwo{CIFAR-100} & \rotvtwo{STL-10} & \rotvtwo{OxfordPets} & \rotvtwo{Flowers102} & \rotvtwo{FGVC} & \rotvtwo{StanfordCars} & \rotvtwo{SUN397} & \rotvtwo{Food101} & \rotvtwo{EuroSAT} & \rotvtwo{DTD} & \rotvtwo{PCAM} & \rotvtwo{Avg. Zero-shot} \\
    \midrule
    \multirow{3}{*}{\rotatebox{90}{Clean}} & TeCoA & 69.30 & 77.40 & 51.90 & 92.90 & 75.00 & 34.90 & 11.00 & 30.10 & 47.60 & 34.40 & \textbf{21.70} & 29.40 & 48.00 & 46.19 \\
    & FARE & 64.40 & \underline{80.60} & \textbf{53.50} & \underline{96.80} & \underline{85.50} & \textbf{57.10} & \textbf{21.20} & \underline{54.90} & \underline{53.10} & \underline{54.80} & 15.30 & \underline{33.70} & \underline{48.00} & \underline{54.54} \\
     & \textbf{(ours) QT-AFT} & 65.90 & \textbf{86.10} & \underline{53.40} & \textbf{96.90} & \textbf{86.00} & \underline{55.00} & \underline{19.50} & \textbf{61.80} & \textbf{53.80} & \textbf{55.80} & \underline{20.00} & \textbf{37.20} & \textbf{60.70} & \textbf{57.18} \\
    \hline

    \multirow{3}{*}{\rotatebox{90}{Adv.}} & TeCoA & 42.00 & 36.70 & \underline{19.80} & 72.20 & \underline{50.80} & \underline{13.30} & \textbf{2.90} & 6.30 & \underline{16.40} & 9.40 & 9.50 & \underline{14.90} & \underline{47.20} & \underline{24.95} \\
    & FARE & 33.20 & \underline{38.00} & 19.00 & \underline{74.30} & \textbf{50.90} & 13.20 & \underline{2.70} & \textbf{11.80} & 15.20 & \underline{10.20} & \textbf{10.40} & 12.70 & \textbf{48.00} & \textbf{25.53} \\
     & \textbf{(ours) QT-AFT} & 32.30 & \textbf{40.60} & \textbf{21.40} & \textbf{74.80} & 45.50 & \textbf{15.00} & 2.50 & \textbf{11.80} & \textbf{17.00} & \textbf{12.30} & \underline{10.30} & \textbf{16.30} & 28.40 & 24.66 \\

    \bottomrule
    \end{tabular}
    }
     
\end{table*}

\subsection{Word Class: Critical Roles of Non-Object Words}

We investigate which types of linguistic information contribute to zero-shot robustness by conducting an input ablation on word classes. 
Our motivation is that, while existing supervised AT methods primarily use class labels, which are often object names, we aim to explore the role of non-object words in enhancing zero-shot robustness.
Specifically, we modify captions by selectively removing certain types of words (e.g., nouns) and evaluate the impact. We use the NLTK toolkit~\footnote{https://www.nltk.org/} to obtain the part-of-speech (POS) tag of each word. 
The full results are shown in Tab.~\ref{tab:abl-text-type}. 
See Appendix~\ref{sec:caption-example-word-analysis} for examples of input ablation captions used in the word class analysis.


\textbf{``Nouns-only'' does not necessarily improve robustness on zero-shot object centric datasets.}
Nouns identify object categories seen during training, and thus restricting captions to only include nouns (``Nouns-only'') might be expected to benefit object-centric datasets.
However, the robustness actually degraded on most zero-shot datasets.
We hypothesize that this is because ``Nouns-only'' captions focus on objects aligned with the training distribution, but lack the descriptive richness necessary to generalize to unseen classes.

\textbf{Adjectives and adverbs contribute to zero-shot robustness by capturing descriptive properties beyond object identity.}
Interestingly, removing adjectives and adverbs (``No adj./adv.'') have negative impact on zero-shot robustness, causing an average degradation of 1.5\%, even reducing performance on object-centric datasets like STL-10, Caltech101, and Caltech256.
We assume that adjectives and adverbs capture descriptive attributes (e.g., color, shape, size), which are transferable across classes and help generalize to unseen categories, improving model performance in zero-shot settings.

\textbf{Removing nouns generally degrades performance but improves robustness in certain tasks where class labels are described using adjectives.}
By removing nouns (``No nouns''), we observe robustness degradation in 14 out of 16 datasets. However, surprisingly, on the texture classification task (DTD), robustness improved from 14\% to 20\%.
This suggests that while nouns are essential for object-centric tasks, their removal enhances robustness in tasks like texture classification, where labels are adjective-based. This highlights the task-dependent nature of language-guided AT.

\textbf{Function words contribute to robustness.}
Function words, such as prepositions (e.g., on, under, next to), conjunctions (e.g., and, or, but), and articles (e.g., the, a), play a critical role in conveying spatial and contextual relationships, which are essential for understanding complex scenes. 
By removing function words (``No function words''), we observe 1\% decrease in both robustness and accuracy.
Their contribution suggests that relational cues help the model capture scene-level semantics and reduce reliance on isolated object identity, leading to more robust and holistic image understanding under adversarial conditions.

\textbf{Word order matters for robustness.}
By shuffling the words and breaking their order (``Shuffle words''), both robustness and accuracy degrade by around 2\%.
This suggests that the structure of captions is crucial for capturing semantic cues and preserving natural language structure enhances robustness by ensuring accurate semantic alignment during AT.


\section{Ablation Study: CLIP-ViT-L/14 results}
For the model size ablation, we train CLIP-ViT-L/14 on ImageNet and compare with two baseline approaches—TeCoA and FARE—using their publicly available model weights.
Tab.~\ref{tab:abl-clip-l14} demonstrates that our proposed method remains effective, achieving significantly higher zero-shot clean accuracy while maintaining strong adversarial robustness, outperforming baselines on 8 out of 12 datasets.

\section{Conclusion}

In this work, we revisited adversarial fine-tuning for pre-trained vision-language models (VLMs) and highlighted the limitations of existing supervised and unsupervised approaches in achieving zero-shot robustness. While supervised methods based on class labels tend to overfit to training data, unsupervised methods fail to target semantically meaningful aspects of images. To address these limitations, we proposed Quality Text-guided Adversarial Fine-Tuning (QT-AFT), which leverages high-quality image captions to guide adversarial example generation toward semantically diverse and descriptive directions.
Through comprehensive experiments, we demonstrated that our method improves both clean and robust zero-shot performance across diverse datasets. Analyses of word-level contributions further revealed that adjectives, adverbs, function words, and even word order play important roles in improving robustness by encoding fine-grained and relational semantics.

Overall, this work introduces a novel perspective on leveraging language for robust vision, emphasizing the importance of semantic richness in adversarial training. We believe that our findings open up promising directions for future research on robust multimodal learning, a distinct direction from unimodal learning.

\begin{acks}
This work was partially supported by JSPS KAKENHI Grants JP21H04907 and JP24H00732, by JST CREST Grant JPMJCR20D3 including AIP challenge program, by JST AIP Acceleration Grant JPMJCR24U3, and by JST K Program Grant JPMJKP24C2 Japan.
\end{acks}

\clearpage

\bibliographystyle{ACM-Reference-Format}
\bibliography{main}

\clearpage

\onecolumn
\appendix

\section{Implementation Details}
\subsection{Baselines}
In this section, we describe the baseline adversarial fine-tuning methods and provide their implementation details.

\begin{itemize}
    \item \textbf{TeCoA~\cite{mao2022understanding}} conducts text-guided contrastive adversarial training by leveraging text embeddings of class labels to obtain a robust vision encoder. We use the official code provided by the authors~\footnote{\url{https://github.com/cvlab-columbia/ZSRobust4FoundationModel}} However, the original paper is limited to training CLIP with a 2-step PGD adversary, for 10 epochs using the SGD optimizer, and a perturbation size of $\epsilon=1/255$. Subsequently, \citet{schlarmann2024robust} showed that modifying these hyperparameters—specifically using 10-step PGD, training for 2 epochs with the AdamW optimizer—yields better performance. Following their findings, we adopt these revised hyperparameters for TeCoA in all our experiments. We also empirically confirm that this configuration consistently results in improved performance.
    \item \textbf{PMG-AFT~\cite{wang2024pre}} improved TeCoA by incorporating guidance from a pre-trained model. We use the original codes and the hyperparamters proposed by the authors~\footnote{https://github.com/serendipity1122/Pre-trained-Model-Guided-Fine-Tuning-for-Zero-Shot-Adversarial-Robustness}.
    \item \textbf{TGA-ZSR~\cite{yu2025text}} enhanced robustness by introducing an attention-guided mechanism. We use the official code and the hyperparamters provided by the authors~\footnote{https://github.com/zhyblue424/TGA-ZSR}. It is worth noting that the original paper did not conduct training on ImageNet and was limited to Tiny-ImageNet~\cite{le2015tiny}, which consists of 100,000 images across 200 classes (500 images per class) resized to 64$\times$64 resolution. This dataset gap leads to  inferior performance when applying the same setup to ImageNet training.
    \item \textbf{FARE~\cite{schlarmann2024robust}} conducts an unsupervised adversarial training without textual guidance. We use the official code and the hyperparamters provided by the authors~\footnote{https://github.com/chs20/RobustVLM}.
    
\end{itemize}

\subsection{Computational Settings}
We use NVIDIA A100 GPUs for all experiments. CLIP-ViT-B/16 is trained on a single A100 GPU and takes approximately 10 hours to complete training on ImageNet. For the larger CLIP-ViT-L/14 model, we use four A100 GPUs, and training takes approximately 6 days.
To generate ImageNet captions, InternVL-2.5-8B was used with two A100 GPUs and required approximately 10 days. Mini-InternVL-1.5-2B required less time, completing in 3 to 4 days.

\subsection{Additional Evaluation Details}
We conduct our evaluation using AutoAttack~\cite{croce2020reliable}, a standard and reliable benchmark for adversarial robustness, widely acknowledged in image classification tasks (see RobustBench\footnote{\url{https://robustbench.github.io/}}). AutoAttack addresses key limitations of PGD-based evaluations, which rely on fixed step sizes and a single objective function, often leading to unreliable results. In contrast, AutoAttack is step-size free and performs an ensemble of attacks, ensuring a more comprehensive and reliable evaluation. In this work, we use two objective functions within AutoAttack: cross-entropy (CE) loss and Difference of Logits Ratio (DLR) loss.

It is worth noting that the numerical precision (Float16 vs. Float32) has a substantial impact on attack performance. Specifically, evaluations using Float32 yield significantly stronger attacks compared to Float16. Therefore, in contrast to \citet{schlarmann2024robust}, we standardize all evaluation settings to use Float32 for consistency and comparability.

\section{Additional Results}
\subsection{Evaluation on Other Attacks}
\label{sec:pgd-10}
In order to conduct evaluation on full samples of 16 zero-shot datasets, we evaluate robustness against 10-step PGD (PGD-10).
Tab.~\ref{tab:zeroshot-pgd10-full} shows that our method, QT-AFT, remains highly effective against PGD-10, improving zero-shot robustness by an average of 5\% while maintaining high clean accuracy.

We further consider two additional attack types.
Tab.~\ref{tab:zeroshot-l2-pgd} reports results for L2-bounded PGD with $\epsilon=128/255$, and Tab.~\ref{tab:zeroshot-cw} shows results for the CW attack~\cite{carlini2017towards}. 
Our method, QT-AFT, outperforms all baselines in both clean and robust accuracy.

\begin{table*}[h]
    \centering
    \caption{\textbf{Clean accuracy and robust accuracy against PGD-10 ($\epsilon=4/255$)} of CLIP evaluated on zero-shot datasets. }
     \label{tab:zeroshot-pgd10-full}
    \resizebox{\textwidth}{!}{
    \begin{tabular}{cc|c|cccccccccccccccc|c}
    \toprule
    & & & \multicolumn{15}{c}{Zero-shot datasets} & \\ \cmidrule(lr){4-19}
      & Method & \rotvtwo{ImageNet} & \rotvtwo{ImageNet-S} & \rotvtwo{ImageNet-R} & \rotvtwo{CIFAR-10} & \rotvtwo{CIFAR-100} & \rotvtwo{STL-10} & \rotvtwo{Caltech101} & \rotvtwo{Caltech256} & \rotvtwo{OxfordPets} & \rotvtwo{Flowers102} & \rotvtwo{FGVC} & \rotvtwo{StanfordCars} & \rotvtwo{SUN397} & \rotvtwo{Food101} & \rotvtwo{EuroSAT} & \rotvtwo{DTD} & \rotvtwo{PCAM} & \rotvtwo{Avg. Zero-shot} \\
    \midrule
    \multirow{5}{*}{\rotatebox{90}{Clean}} & PMG-AFT & 56.98 & 31.80 & 52.44 & \underline{78.20} & 47.31 & 92.58 & 80.38 & 66.02 & 71.98 & 11.58 & 3.15 & 9.30 & 35.26 & 28.76 & 16.45 & 24.63 & \underline{51.68} & 43.84 \\
& TGA-ZSR & \textbf{70.22} & \textbf{40.65} & \textbf{62.94} & \textbf{87.81} & \textbf{59.34} & \textbf{96.78} & 79.56 & \textbf{79.58} & \textbf{79.89} & \textbf{43.86} & \textbf{15.66} & 35.18 & \textbf{54.04} & \textbf{62.81} & \underline{21.26} & \textbf{32.87} & 48.99 & \textbf{56.33} \\
& TeCoA & \underline{65.47} & 32.93 & 55.14 & 76.56 & 43.75 & 91.02 & 76.56 & 64.58 & \underline{76.17} & 20.05 & 5.86 & 11.72 & 36.16 & 22.03 & 20.81 & 21.88 & \textbf{58.65} & 44.62 \\
& FARE & 53.63 & 36.02 & 55.14 & 70.90 & 44.53 & \underline{93.95} & \underline{83.01} & 72.07 & 74.22 & \underline{40.62} & \underline{14.06} & \textbf{41.02} & 44.89 & \underline{43.98} & \textbf{23.93} & \underline{30.47} & 48.86 & \underline{51.10} \\
& \textbf{(ours) QT-AFT} & 53.14 & \underline{36.70} & \underline{60.90} & 73.24 & \underline{53.32} & 93.75 & \textbf{85.49} & \underline{74.61} & 75.00 & 22.50 & 10.23 & \underline{37.98} & \underline{48.84} & 41.69 & 17.72 & 29.17 & 50.61 & 50.73 \\
  \hline
        
    \multirow{5}{*}{\rotatebox{90}{Adv.}} & PMG-AFT & \underline{31.45} & \underline{17.75} & 28.80 & \textbf{41.96} & \underline{22.06} & \underline{70.97} & \underline{59.20} & 39.99 & \underline{47.02} & 3.94 & 1.08 & 2.03 & 12.85 & 7.00 & 11.18 & 13.09 & 28.77 & 25.48 \\
& TGA-ZSR & 17.26 & 4.57 & 17.53 & 28.43 & 10.72 & 59.32 & 47.22 & 34.67 & 32.71 & 2.26 & 0.00 & 0.06 & 3.35 & \underline{8.48} & 0.06 & 3.51 & 0.00 & 15.81 \\
& TeCoA & \textbf{35.08} & 17.70 & \underline{29.75} & 40.04 & 17.97 & 67.38 & 54.88 & 37.37 & 42.58 & 5.99 & \underline{2.34} & 3.71 & \underline{13.74} & 6.02 & \textbf{13.07} & 11.72 & 21.88 & 24.13 \\
& FARE & 21.41 & 16.33 & 24.74 & 39.84 & 19.53 & 68.55 & 56.25 & \underline{40.23} & 32.42 & \underline{10.68} & 1.56 & \underline{4.49} & 11.95 & 7.50 & 7.17 & \textbf{18.75} & \textbf{48.80} & \underline{25.55} \\
& \textbf{(ours) QT-AFT} & 29.38 & \textbf{21.27} & \textbf{30.78} & \underline{41.60} & \textbf{23.44} & \textbf{72.75} & \textbf{65.18} & \textbf{48.05} & \textbf{48.44} & \textbf{11.88} & \textbf{3.41} & \textbf{11.30} & \textbf{20.86} & \textbf{9.97} & \underline{11.25} & \underline{17.71} & \underline{44.42} & \textbf{30.14} \\

    \bottomrule
    \end{tabular}
    }
\end{table*}

\begin{table*}[h]
    \centering
    \caption{\textbf{Clean accuracy and robust accuracy against L2-PGD ($\epsilon=128/255$)} of CLIP evaluated on zero-shot datasets. }
     \label{tab:zeroshot-l2-pgd}
    \resizebox{\textwidth}{!}{
    \begin{tabular}{cc|c|cccccccccccccccc|c}
    \toprule
    & & & \multicolumn{15}{c}{Zero-shot datasets} & \\ \cmidrule(lr){4-19}
      & Method & \rotvtwo{ImageNet} & \rotvtwo{ImageNet-S} & \rotvtwo{ImageNet-R} & \rotvtwo{CIFAR-10} & \rotvtwo{CIFAR-100} & \rotvtwo{STL-10} & \rotvtwo{Caltech101} & \rotvtwo{Caltech256} & \rotvtwo{OxfordPets} & \rotvtwo{Flowers102} & \rotvtwo{FGVC} & \rotvtwo{StanfordCars} & \rotvtwo{SUN397} & \rotvtwo{Food101} & \rotvtwo{EuroSAT} & \rotvtwo{DTD} & \rotvtwo{PCAM} & \rotvtwo{Avg. Zero-shot} \\
    \midrule
    \multirow{5}{*}{\rotatebox{90}{Clean}} & PMG-AFT  & 55.60 & 31.70 & 50.90 & \underline{76.60} & 45.90 & 92.50 & 77.70 & 67.50 & 67.10 & 9.90 & 2.90 & 8.60 & 33.40 & 27.90 & \textbf{23.50} & 24.80 & 48.00 & 43.79 \\
& TGA-ZSR & \textbf{69.60} & \textbf{38.50} & \textbf{62.40} & \textbf{87.90} & \textbf{56.40} & \textbf{96.90} & 78.60 & \textbf{80.40} & \textbf{78.90} & \textbf{44.80} & \textbf{16.30} & 33.30 & \textbf{52.30} & \textbf{64.00} & \underline{22.50} & \textbf{32.80} & 46.90 & \textbf{56.62} \\
& TeCoA & \underline{63.30} & 31.80 & 51.90 & 75.20 & 39.10 & 91.70 & 74.70 & 66.40 & 71.80 & 19.50 & 6.90 & 12.60 & 35.90 & 20.90 & 17.00 & 21.40 & \textbf{57.90} & 44.59 \\
& FARE& 50.60 & 35.60 & 57.00 & 64.50 & 47.30 & 91.80 & \underline{80.50} & 74.40 & \underline{76.40} & \underline{39.10} & \underline{13.50} & \textbf{39.50} & 42.90 & \underline{44.30} & 21.90 & 27.00 & 48.00 & 50.25 \\
& \textbf{(ours) QT-AFT} & 53.70 & \underline{36.20} & \underline{58.30} & 71.70 & \underline{49.20} & \underline{93.30} & \textbf{81.80} & \underline{76.10} & 74.00 & 33.20 & 13.30 & \underline{37.70} & \underline{48.30} & 43.30 & 17.20 & \underline{29.50} & \underline{48.00} & \underline{50.87} \\

  \hline
        
    \multirow{5}{*}{\rotatebox{90}{Adv.}} & PMG-AFT  & \underline{52.00} & 29.40 & 47.50 & \textbf{46.80} & \textbf{23.40} & \underline{88.00} & 74.90 & 63.30 & 63.60 & 8.30 & 2.20 & 7.00 & 29.70 & 24.00 & \textbf{21.30} & 23.20 & 48.00 & 38.39 \\
& TGA-ZSR & 15.00 & 6.40 & 18.60 & 8.40 & 3.10 & 62.20 & 46.60 & 35.80 & 24.80 & 6.10 & 0.40 & 1.20 & 8.10 & 13.00 & 0.10 & 9.00 & 1.40 & 15.31 \\
& TeCoA & \textbf{60.50} & 29.90 & 48.10 & \underline{44.20} & 21.20 & 86.90 & 72.00 & 63.80 & 67.30 & 17.90 & 5.50 & 11.60 & 32.60 & 18.80 & 16.20 & 20.40 & \textbf{54.10} & 39.47 \\
& FARE& 46.50 & \underline{31.90} & \underline{50.80} & 39.30 & 21.20 & 86.80 & \underline{78.50} & \underline{69.90} & \textbf{70.30} & \textbf{34.10} & \textbf{10.50} & \textbf{32.60} & \underline{37.70} & \underline{38.10} & \underline{19.80} & \underline{25.10} & \underline{48.00} & \underline{43.59} \\
& \textbf{(ours) QT-AFT} & 49.00 & \textbf{32.40} & \textbf{52.30} & 42.70 & 23.40 & \textbf{88.70} & \textbf{79.10} & \textbf{71.30} & \underline{69.10} & \underline{29.50} & \underline{10.20} & \underline{30.10} & \textbf{42.90} & \textbf{38.60} & 15.90 & \textbf{27.30} & 47.90 & \textbf{44.14} \\

    \bottomrule
    \end{tabular}
    }
\end{table*}

\begin{table*}[h]
    \centering
    \caption{\textbf{Clean accuracy and robust accuracy against CW-Attack ($\epsilon=4/255$ in $\ell_{\infty}$-norm)} of CLIP evaluated on zero-shot datasets. }
     \label{tab:zeroshot-cw}
    \resizebox{\textwidth}{!}{
    \begin{tabular}{cc|c|cccccccccccccccc|c}
    \toprule
    & & & \multicolumn{15}{c}{Zero-shot datasets} & \\ \cmidrule(lr){4-19}
      & Method & \rotvtwo{ImageNet} & \rotvtwo{ImageNet-S} & \rotvtwo{ImageNet-R} & \rotvtwo{CIFAR-10} & \rotvtwo{CIFAR-100} & \rotvtwo{STL-10} & \rotvtwo{Caltech101} & \rotvtwo{Caltech256} & \rotvtwo{OxfordPets} & \rotvtwo{Flowers102} & \rotvtwo{FGVC} & \rotvtwo{StanfordCars} & \rotvtwo{SUN397} & \rotvtwo{Food101} & \rotvtwo{EuroSAT} & \rotvtwo{DTD} & \rotvtwo{PCAM} & \rotvtwo{Avg. Zero-shot} \\
    \midrule
    \multirow{5}{*}{\rotatebox{90}{Clean}} & PMG-AFT  & 55.60 & 31.70 & 50.90 & \underline{76.60} & 45.90 & 92.50 & 77.70 & 67.50 & 67.10 & 9.90 & 2.90 & 8.60 & 33.40 & 27.90 & \textbf{23.50} & 24.80 & 48.00 & 43.79 \\
& TGA-ZSR & \textbf{69.60} & \textbf{38.50} & \textbf{62.40} & \textbf{87.90} & \textbf{56.40} & \textbf{96.90} & 78.60 & \textbf{80.40} & \textbf{78.90} & \textbf{44.80} & \textbf{16.30} & 33.30 & \textbf{52.30} & \textbf{64.00} & \underline{22.50} & \textbf{32.80} & 46.90 & \textbf{56.62} \\
& TeCoA & \underline{63.30} & 31.80 & 51.90 & 75.20 & 39.10 & 91.70 & 74.70 & 66.40 & 71.80 & 19.50 & 6.90 & 12.60 & 35.90 & 20.90 & 17.00 & 21.40 & \textbf{57.90} & 44.59 \\
& FARE& 50.60 & 35.60 & 57.00 & 64.50 & 47.30 & 91.80 & \underline{80.50} & 74.40 & \underline{76.40} & \underline{39.10} & \underline{13.50} & \textbf{39.50} & 42.90 & \underline{44.30} & 21.90 & 27.00 & 48.00 & 50.25 \\
& \textbf{(ours) QT-AFT} & 53.70 & \underline{36.20} & \underline{58.30} & 71.70 & \underline{49.20} & \underline{93.30} & \textbf{81.80} & \underline{76.10} & 74.00 & 33.20 & 13.30 & \underline{37.70} & \underline{48.30} & 43.30 & 17.20 & \underline{29.50} & \underline{48.00} & \underline{50.87} \\

  \hline
        
    \multirow{5}{*}{\rotatebox{90}{Adv.}} & PMG-AFT  & \underline{31.60} & 15.10 & \underline{26.10} & \textbf{37.80} & \textbf{18.90} & \textbf{71.10} & \underline{56.80} & 36.80 & \underline{40.30} & 3.20 & 0.20 & 1.40 & 11.40 & 6.20 & 3.70 & 10.40 & \underline{47.70} & \underline{24.63} \\
& TGA-ZSR & 0.20 & 0.20 & 0.00 & 0.00 & 0.10 & 0.00 & 0.20 & 0.00 & 0.00 & 0.00 & 0.00 & 2.10 & 0.00 & 0.00 & 0.00 & 0.00 & 0.00 & 0.16 \\
& TeCoA & \textbf{34.40} & 15.00 & \textbf{27.00} & 33.50 & 17.90 & \underline{69.00} & 50.60 & \underline{37.80} & \textbf{40.80} & 5.90 & 1.30 & 3.40 & \underline{12.40} & 6.30 & \underline{10.30} & 10.30 & 20.30 & 23.31 \\
& FARE& 21.60 & \underline{15.30} & 22.30 & 31.20 & 15.60 & 63.50 & 54.50 & 37.40 & 33.60 & \underline{9.20} & \underline{2.30} & \textbf{6.30} & 11.80 & \underline{8.60} & 4.00 & \underline{13.30} & \textbf{48.00} & 23.44 \\
& \textbf{(ours) QT-AFT} & 24.00 & \textbf{16.20} & 24.00 & \underline{33.50} & \underline{18.80} & 67.40 & \textbf{57.30} & \textbf{40.50} & 35.50 & \textbf{10.60} & \textbf{3.20} & 6.30 & \textbf{14.50} & \textbf{8.70} & \textbf{11.20} & \textbf{14.60} & 42.30 & \textbf{25.21} \\

    \bottomrule
    \end{tabular}
    }
\end{table*}

\subsection{QT-AFT as a Vision Encoder for Large Vision-Language Models}
In the main paper, we focused on fine-tuning the vision encoder of CLIP to improve robustness. Beyond CLIP, Large Vision-Language Models (LVLMs) such as LLaVA~\cite{liu2023llava} and OpenFlamingo~\cite{awadalla2023openflamingo} are increasingly deployed in real-world applications, making them susceptible to adversarial image attacks. Since both LLaVA and OpenFlamingo rely on the CLIP vision encoder, we can enhance the robustness of these LVLMs by substituting their vision encoder with the QT-AFT-trained robust encoder, without modifying their language models.

To evaluate this approach, we replaced the vision encoder in LLaVA and OpenFlamingo-9B with QT-AFT and conducted experiments on COCO and Flickr30k for image captioning, as well as TextVQA and VQAv2 for visual question answering.
Table~\ref{tab:flamingo_llava_results} shows that QT-AFT achieved robustness and clean accuracy comparable to FARE. QT-AFT outperforms TeCoA, likely because it avoids class-label overfitting. 
We also note that additional gains are expected by fine-tuning the MLP projector, as QT-AFT does not impose strong constraints on embedding shifts (unlike FARE).

\begin{table*}[h]
\centering
\caption{Clean and Robust Accuracy for Open Flamingo-9B and LLaVA across datasets. We replaced the vision encoder with robust CLIP vision encoder.}
\begin{tabular}{c lcccccccc}
\toprule
\textbf{} & \textbf{Method} 
& \multicolumn{4}{c}{\textbf{Open Flamingo-9B}} 
& \multicolumn{4}{c}{\textbf{LLaVA}} \\
\cmidrule(lr){3-6} \cmidrule(lr){7-10}
& & \textbf{COCO} & \textbf{Flickr30k} & \textbf{TextVQA} & \textbf{VQAv2} 
& \textbf{COCO} & \textbf{Flickr30k} & \textbf{TextVQA} & \textbf{VQAv2} \\
\midrule
\multirow{4}{*}{\rotatebox{90}{Clean}} 
& (Pretrained) & 88.48 & 61.43 & 18.96 & 45.48 & 122.38 & 79.25 & 37.26 & 72.78 \\ \cdashline{2-10}
& TeCoA        & 71.58 & 42.98 & 11.42 & 44.48 & 96.19  & 52.11 & 20.12 & 62.16 \\
& FARE         & \textit{78.88} & \textbf{54.76} & \textbf{17.22} & \textit{44.80} 
               & \textbf{106.04} & \textit{64.93} & \textit{26.90} & \textit{65.76} \\
& QT-AFT       & \textbf{82.34} & \textit{51.59} & \textit{15.84} & \textbf{45.52} 
               & \textit{105.76} & \textbf{65.81} & \textbf{27.00} & \textbf{66.40} \\
\midrule
\multirow{4}{*}{\rotatebox{90}{Robust}} 
& (Pretrained) & 1.22  & 0.47  & 0.00  & 0.68  & 2.78   & 0.96  & 0.00  & 0.00  \\ \cdashline{2-10}
& TeCoA        & 22.22 & \textit{8.89} & \textit{2.48} & \textbf{22.04} 
               & 34.47 & 19.51         & \textit{9.34} & \textit{30.20} \\
& FARE         & \textbf{23.99} & 10.14 & \textbf{2.58} & \textit{21.28} 
               & \textbf{42.06} & \textbf{23.02} & \textbf{10.32} & 29.88 \\
& QT-AFT       & \textit{23.47} & \textbf{11.55} & 2.28  & 21.08 
               & \textit{39.16} & \textit{22.51} & 8.30  & \textbf{30.34} \\
\bottomrule
\end{tabular}
\label{tab:flamingo_llava_results}
\end{table*}

\subsection{Additional Caption Comparison}
To assess the generalization of QT-AFT to other captioning models, we present additional results using Qwen2.5-VL-3B-Instruct~\footnote{https://huggingface.co/Qwen/Qwen2.5-VL-3B-Instruct}~\cite{Qwen2VL}.
This model is relatively small, with 3B parameters, compared to the captioning model used in the main paper, InternVL-2.5-8B. Caption generation for ImageNet was completed in 3 days on a single A100 GPU, producing rich captions of approximately 50 words.

Despite its smaller size, QT-AFT with Qwen-3B achieved performance comparable to QT-AFT with InternVL-8B, demonstrating strong generalization.
This highlights that our approach is promising as VLMs continue become more efficient and effective.

\begin{table}[h]
    \centering
    \caption{\textbf{Caption Quality Analysis: Image Captioning Model.} Clean accuracy and AutoAttack robustness ($\epsilon=4/255$) of CLIP trained on ImageNet with QT-AFT, using captions generated by different VL models. Using a smaller VL model to generate captions also achieves state-of-the-art performance; however, leveraging a stronger captioning model achives the best results.}
     \label{tab:abl-captioning-model}
    \resizebox{\textwidth}{!}{
    \begin{tabular}{cc|c|cccccccccccccccc|c}
    \toprule
    & & & \multicolumn{16}{c}{Zero-shot datasets} & \\ \cmidrule(lr){4-19}
      & Method & \rotvtwo{ImageNet} & \rotvtwo{ImageNet-S} & \rotvtwo{ImageNet-R} & \rotvtwo{CIFAR-10} & \rotvtwo{CIFAR-100} & \rotvtwo{STL-10} & \rotvtwo{Caltech101} & \rotvtwo{Caltech256} & \rotvtwo{OxfordPets} & \rotvtwo{Flowers102} & \rotvtwo{FGVC} & \rotvtwo{StanfordCars} & \rotvtwo{SUN397} & \rotvtwo{Food101} & \rotvtwo{EuroSAT} & \rotvtwo{DTD} & \rotvtwo{PCAM} & \rotvtwo{Avg. Zero-shot} \\
    \midrule
    \multirow{2}{*}{\rotatebox{90}{Clean}} & \textbf{QT-AFT w/ Mini-IntVL-1.5-2B} & \textbf{52.1} & \underline{36.2} & \textbf{57.8} & \textbf{73.8} & \textbf{49.0} & \underline{93.8} & \textbf{82.8} & \textbf{75.6} & \underline{75.0} & \textbf{33.8} & \underline{11.5} & \underline{38.0} & \underline{48.3} & \textbf{44.8} & \textbf{19.6} & \underline{21.9} & \textbf{52.0} & \underline{50.9} \\
& \textbf{QT-AFT w/ IntVL-2.5-8B (default)} & \underline{51.9} & \textbf{38.5} & \underline{56.9} & \underline{70.9} & \underline{48.6} & \textbf{95.8} & \underline{81.9} & \underline{73.4} & \textbf{80.7} & \underline{30.6} & \textbf{12.5} & \textbf{40.1} & \textbf{51.7} & \underline{44.2} & \underline{19.2} & \textbf{29.2} & \underline{51.1} & \textbf{51.6} \\
& \textbf{QT-AFT w/ Qwen-3B} & \textbf{52.2} & \textbf{39.1} & \textbf{61.7} & \underline{71.1} & 47.5 & 92.3 & \textbf{84.2} & \underline{73.6} & \textbf{83.9} & 30.0 & \textbf{16.5} & \textbf{41.1} & 44.7 & \textbf{45.1} & \textbf{20.4} & \textbf{34.4} & 51.0 & \textbf{52.3} \\

 \hline

    \multirow{2}{*}{\rotatebox{90}{Adv.}} 
& \textbf{QT-AFT w/ Mini-IntVL-1.5-2B} & \textbf{20.5} & \underline{15.2} & \underline{22.4} & \underline{32.4} & \underline{19.1} & \underline{67.1} & \underline{54.5} & \textbf{42.0} & \textbf{36.5} & \textbf{10.9} & \underline{1.0} & \underline{5.3} & \underline{13.0} & \textbf{8.5} & \underline{12.4} & \underline{6.2} & \textbf{49.3} & \underline{24.7} \\
& \textbf{QT-AFT w/ IntVL-2.5-8B (default)} & \underline{19.6} & \textbf{17.6} & \textbf{25.2} & \textbf{33.2} & \textbf{20.9} & \textbf{69.0} & \textbf{58.9} & \underline{40.6} & \textbf{36.5} & \underline{9.7} & \textbf{2.3} & \textbf{5.8} & \textbf{14.2} & \underline{7.7} & \textbf{12.6} & \textbf{14.6} & \underline{44.1} & \textbf{25.8} \\
& \textbf{QT-AFT w/ Qwen-3B} & \textbf{21.2} & \textbf{19.5} & \underline{23.1} & 31.6 & 15.2 & 62.8 & \underline{56.9} & \underline{41.7} & \textbf{39.6} & 5.9 & 0.57 & \textbf{8.4} & \underline{14.1} & \underline{8.3} & \textbf{13.8} & \textbf{17.7} & \underline{48.3} & \underline{25.5} \\
    \bottomrule
    \end{tabular}
    }
\end{table}

\subsection{Hyperparameter $\lambda$}
We conducted a sweep over the hyperparameter $\lambda$ in Eq.~\ref{eq:ours}, which determines the balance between unsupervised and caption-guided loss.
Table~\ref{tab:hparam_lambda} shows that the results are robust, with $\lambda=10.0$ performing best.  

\begin{table}[h]
    \centering
    \caption{Effect of $\lambda$ on clean and adversarial accuracy.}
    \begin{tabular}{lcc}
        \toprule
         & \textbf{Clean (Avg.)} & \textbf{Adv (Avg.)} \\
        \midrule
        QT-AFT ($\lambda=1.0$)         & 51.3 & 24.7 \\
        QT-AFT ($\lambda=5.0$)         & 51.0 & 24.9 \\
        \textbf{QT-AFT ($\lambda=10.0$, default)} & \textbf{51.6} & \textbf{25.8} \\
        QT-AFT ($\lambda=15.0$)        & 50.3 & 24.4 \\
        \bottomrule
    \end{tabular}
    \label{tab:hparam_lambda}
\end{table}

\clearpage

\section{Examples of Captions}
\subsection{Image Captioning Model}
\label{sec:image-captioning-examples}

\begin{figure*}[h]
  \centering
  \includegraphics[width=0.7\linewidth]{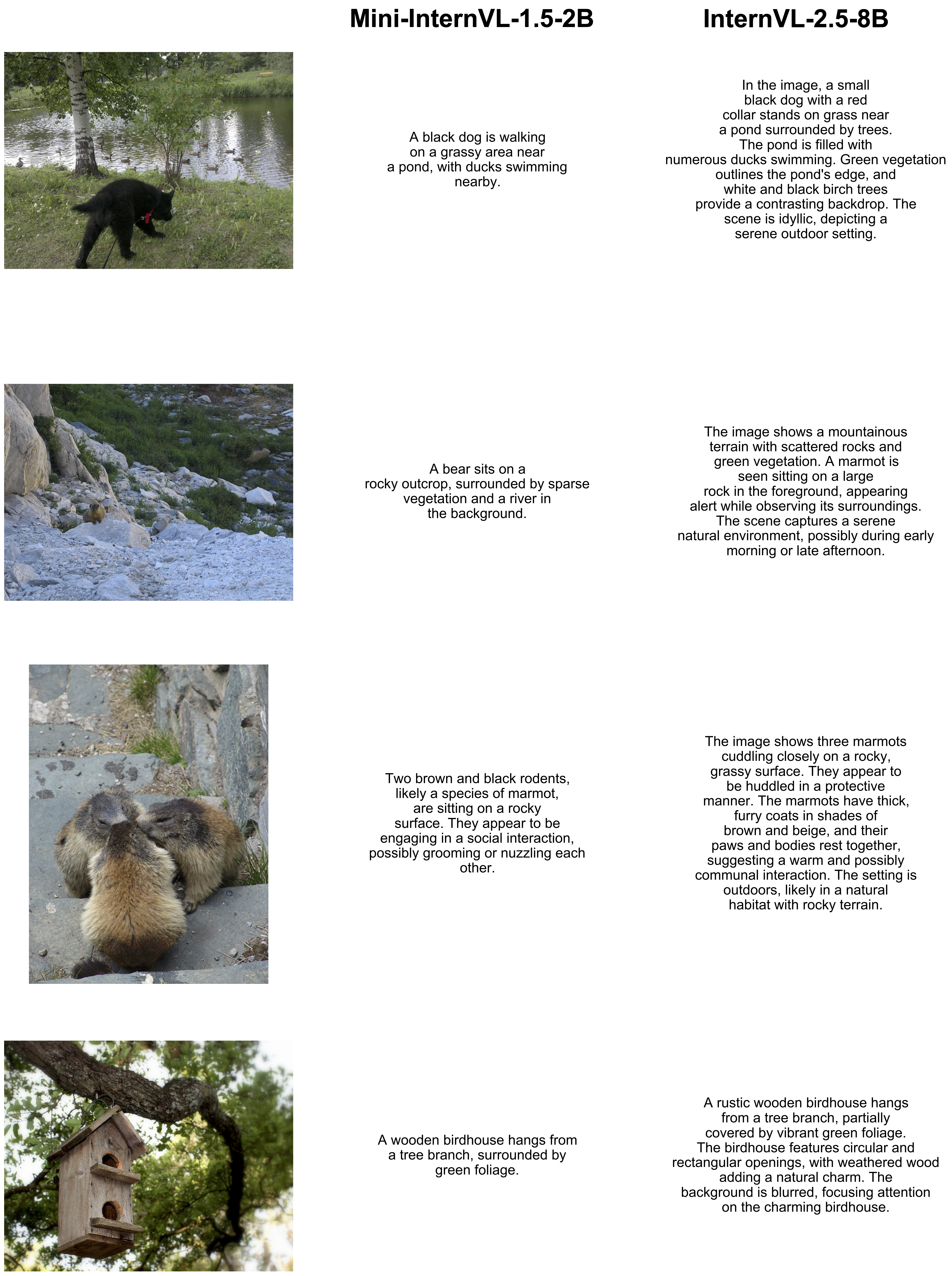}
  \caption{Image caption comparison between Mini-InternVL-1.5-2B and InternVL-2.5-8B. By default, QT-AFT uses captions generated by InternVL-2.5-8B. Compared to the relatively short captions produced by Mini-InternVL-1.5-2B, InternVL-2.5-8B generates more detailed and descriptive captions, offering richer semantics that are more effective for guiding adversarial perturbations during QT-AFT training.}
  \label{fig:caption-comparison}
\end{figure*}

Figure~\ref{fig:caption-comparison} presents example captions generated by Mini-InternVL-1.5-2B and InternVL-2.5-8B. 

\clearpage

\subsection{Word Class}
\label{sec:caption-example-word-analysis}

\begin{figure*}[h]
  \centering
  \includegraphics[width=0.25\linewidth]{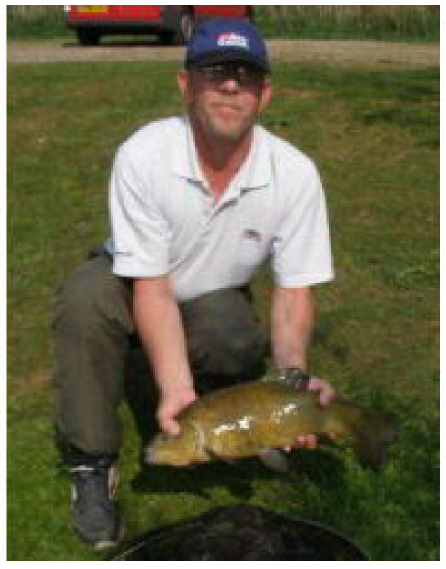}
  \caption{An example image from ImageNet.}
  \label{fig:example-image}
\end{figure*}

Here, we provide example captions used in Sec.\ref{subsec:analysis} for the image shown in Fig.\ref{fig:example-image}.

\begin{tcolorbox}[colframe=blue, colback=white, coltitle=black, width=\textwidth, title=Original caption (generated by InternVL-2.5-8B):, coltitle=white]
A man kneels on grass, holding a large fish above a grassy field. He wears sunglasses and a cap, dressed in a white polo shirt and camouflage pants, sporting grey sneakers. In front of him, a net or container rests on the ground, likely used for the catch. The green hues of the grass contrast with the man's light clothing and the dark fish.
\end{tcolorbox}

\begin{tcolorbox}[colframe=black, colback=white, coltitle=black, width=\textwidth, title=Nouns-Only (only using nouns), coltitle=white]
man,grass,fish,field,sunglasses,cap,polo,shirt,camouflage,pants,sneakers,front,container,rests,ground,catch,hues,contrast,clothing
\end{tcolorbox}

\begin{tcolorbox}[colframe=black, colback=white, coltitle=black, width=\textwidth, title=No adj./adv. (removing adjectives and adverbs), coltitle=white]
A man kneels on grass , holding a fish above a field . He wears sunglasses and a cap , dressed in a polo shirt and camouflage pants , sporting sneakers . In front of him , a or container rests on the ground , used for the catch . The hues of the grass contrast with the man 's clothing and the fish .
\end{tcolorbox}

\begin{tcolorbox}[colframe=black, colback=white, coltitle=black, width=\textwidth, title=No nouns (removing nouns), coltitle=white]
A kneels on , holding a large above a grassy . He wears and a , dressed in a white and , sporting grey . In of him , a net or on the , likely used for the . The green of the with the 's light and the dark .
\end{tcolorbox}

\begin{tcolorbox}[colframe=black, colback=white, coltitle=black, width=\textwidth, title=No function words (removing function words), coltitle=white]
man kneels grass holding large fish grassy field wears sunglasses cap dressed white polo shirt camouflage pants sporting grey sneakers front net container rests ground likely used catch green hues grass contrast man light clothing dark fish
\end{tcolorbox}

\begin{tcolorbox}[colframe=black, colback=white, coltitle=black, width=\textwidth, title=Shuffle words, coltitle=white]
with a The of polo grey the , him sneakers hues and , holding camouflage kneels grass of container a a catch green , dark the net used and for field , in the clothing large cap likely front light sporting above or on and wears pants sunglasses contrast man white ground the grassy a fish man shirt . , A . He . the a dressed fish 's In rests grass on .
\end{tcolorbox}

\begin{tcolorbox}[colframe=orange, colback=white, coltitle=black, width=\textwidth, title=Caption generated by Mini-InternVL-1.5-2B, coltitle=white]
A man in a white polo shirt and cap is kneeling on grass holding a fish.
\end{tcolorbox}

\end{document}